\documentclass[preprint,12pt]{article}

\usepackage{amssymb}
\usepackage{amsthm}
\usepackage{latexsym}
\usepackage[T1]{fontenc}
\usepackage[latin1]{inputenc}
\usepackage[english]{babel}
\usepackage{amsfonts}
\usepackage{amsmath}
\usepackage{float}
\usepackage{subfig}
\usepackage{graphicx}
\usepackage{rotating}
\usepackage[table]{xcolor}
\usepackage{multirow}
\usepackage[ruled]{algorithm2e}
\usepackage{titlesec}
\usepackage {tikz}
\usepackage{thmtools,thm-restate}
\usepackage{authblk}
\usetikzlibrary {positioning}
\titleformat{\paragraph}
{\normalfont\normalsize\bfseries}{\theparagraph}{1em}{}
\titlespacing*{\paragraph}
{0pt}{3.25ex plus 1ex minus .2ex}{1.5ex plus .2ex}

\tikzset{
  font={\fontsize{10pt}{12}\selectfont}}
  
\definecolor{darkGreen}{rgb}{0,0,0.9}
\definecolor{darkRed}{rgb}{0.99,0,0}
\definecolor {processblue}{cmyk}{0.96,0,0,0}

\tikzstyle{C1} = [rectangle, rounded corners, minimum width=0.7cm, minimum height=0.4cm,text centered,text width=0.7cm, draw=black, fill=green!30]
\tikzstyle{C1G} = [rectangle, rounded corners, minimum width=0.8cm, minimum height=0.4cm,text centered,text width=0.8cm, draw=black, fill=green!30]
\tikzstyle{C2} = [rectangle, rounded corners, minimum width=2.3cm, minimum height=0.4cm,text centered,text width=2.3cm, draw=black, fill=green!30]
\tikzstyle{CM} = [rectangle, rounded corners, minimum width=0.5cm, minimum height=0.2cm,text centered,text width=0.5cm, draw=white, fill=white!30]
\tikzstyle{C22} = [rectangle, rounded corners, minimum width=2.7cm, minimum height=0.2cm,text centered,text width=2.3cm, draw=white, fill=white!30]
\tikzstyle{C4G} = [rectangle, rounded corners, minimum width=1.45cm, minimum height=0.4cm,text centered,text width=1.45cm, draw=black, fill=green!30]
\tikzstyle{C44Y} = [rectangle, rounded corners, minimum width=1.45cm, minimum height=0.4cm,text centered,text width=1.45cm, draw=black, fill=yellow!30]
\tikzstyle{C44B} = [rectangle, rounded corners, minimum width=1.45cm, minimum height=0.4cm,text centered,text width=1.45cm, draw=black, fill=blue!30]
\tikzstyle{C44R} = [rectangle, rounded corners, minimum width=1.45cm, minimum height=0.4cm,text centered,text width=1.45cm, draw=black, fill=red!30]
\tikzstyle{C44P} = [rectangle, rounded corners, minimum width=1.45cm, minimum height=0.4cm,text centered,text width=1.45cm, draw=black, fill=purple!30]
\tikzstyle{C44O} = [rectangle, rounded corners, minimum width=1.45cm, minimum height=0.4cm,text centered,text width=1.45cm, draw=black, fill=orange!30]
\tikzstyle{C44T} = [rectangle, rounded corners, minimum width=1.45cm, minimum height=0.4cm,text centered,text width=1.45cm, draw=black, fill=teal!30]
\tikzstyle{C44LI} = [rectangle, rounded corners, minimum width=1.45cm, minimum height=0.4cm,text centered,text width=1.45cm, draw=black, fill=olive!30]
\tikzstyle{C44LG} = [rectangle, rounded corners, minimum width=1.45cm, minimum height=0.4cm,text centered,text width=1.45cm, draw=black, fill=lightgray!30]

\tikzstyle{C11} = [rectangle, rounded corners, minimum width=0.7cm, minimum height=0.4cm,text centered,text width=0.7cm, draw=black, fill=yellow!30]
\tikzstyle{C1O} = [rectangle, rounded corners, minimum width=0.8cm, minimum height=0.4cm,text centered,text width=0.8cm, draw=black, fill=yellow!30]
\tikzstyle{C2O} = [rectangle, rounded corners, minimum width=2.3cm, minimum height=0.4cm,text centered,text width=2.3cm, draw=black, fill=yellow!30]
\tikzstyle{C3O} = [rectangle, rounded corners, minimum width=1.8cm, minimum height=0.4cm,text centered,text width=1.8cm, draw=black, fill=yellow!30]

\tikzstyle{CMW} = [rectangle, rounded corners, minimum width=0.7cm, minimum height=0.4cm,text centered,text width=0.7cm, draw=white, fill=white!30]
\tikzstyle{C11O} = [rectangle, rounded corners, minimum width=0.7cm, minimum height=0.4cm,text centered,text width=0.7cm, draw=black, fill=orange!30]
\tikzstyle{C11B} = [rectangle, rounded corners, minimum width=0.7cm, minimum height=0.4cm,text centered,text width=0.7cm, draw=black, fill=blue!30]
\tikzstyle{C11R} = [rectangle, rounded corners, minimum width=0.7cm, minimum height=0.4cm,text centered,text width=0.7cm, draw=black, fill=red!30]
\tikzstyle{C11R2} = [rectangle, rounded corners, minimum width=0.7cm, minimum height=0.4cm,text centered,text width=0.7cm, draw=darkGreen, fill=red!30]
\tikzstyle{C11P} = [rectangle, rounded corners, minimum width=0.7cm, minimum height=0.4cm,text centered,text width=0.7cm, draw=black, fill=purple!30]
\tikzstyle{C11BR} = [rectangle, rounded corners, minimum width=0.7cm, minimum height=0.4cm,text centered,text width=0.7cm, draw=black, fill=brown!30]
\tikzstyle{C11T} = [rectangle, rounded corners, minimum width=0.7cm, minimum height=0.4cm,text centered,text width=0.7cm, draw=black, fill=teal!30]
\tikzstyle{C11LI} = [rectangle, rounded corners, minimum width=0.7cm, minimum height=0.4cm,text centered,text width=0.7cm, draw=black, fill=olive!30]
\tikzstyle{C11LI2} = [rectangle, rounded corners, minimum width=0.7cm, minimum height=0.4cm,text centered,text width=0.7cm, draw=darkGreen, fill=olive!30]
\tikzstyle{C11LG} = [rectangle, rounded corners, minimum width=0.7cm, minimum height=0.4cm,text centered,text width=0.7cm, draw=black, fill=lightgray!30]
\tikzstyle{C1OB} = [rectangle, rounded corners, minimum width=0.8cm, minimum height=0.4cm,text centered,text width=0.8cm, draw=black, fill=blue!30]
\tikzstyle{C1OR} = [rectangle, rounded corners, minimum width=0.8cm, minimum height=0.4cm,text centered,text width=0.8cm, draw=black, fill=red!30]
\tikzstyle{C1OP} = [rectangle, rounded corners, minimum width=1.1cm, minimum height=0.4cm,text centered,text width=1.1cm, draw=black, fill=purple!30]
\tikzstyle{C1OBR} = [rectangle, rounded corners, minimum width=1.1cm, minimum height=0.4cm,text centered,text width=1.1cm, draw=black, fill=brown!30]
\tikzstyle{C1OO} = [rectangle, rounded corners, minimum width=1.1cm, minimum height=0.4cm,text centered,text width=1.1cm, draw=black, fill=orange!30]
\tikzstyle{C1OT} = [rectangle, rounded corners, minimum width=0.8cm, minimum height=0.4cm,text centered,text width=0.8cm, draw=black, fill=teal!30]
\tikzstyle{C1OLI} = [rectangle, rounded corners, minimum width=0.8cm, minimum height=0.4cm,text centered,text width=0.8cm, draw=black, fill=olive!30]
\tikzstyle{C1OLG} = [rectangle, rounded corners, minimum width=1.1cm, minimum height=0.4cm,text centered,text width=1.1cm, draw=black, fill=lightgray!30]

\tikzstyle{CL} = [rectangle, rounded corners, minimum width=2.1cm, minimum height=0.4cm,text centered,text width=2.1cm, draw=black, fill=orange!30]

\tikzstyle{CLG} = [rectangle, rounded corners, minimum width=1.8cm, minimum height=0.4cm,text centered,text width=1.8cm, draw=black, fill=green!30]

\tikzstyle{C1B} = [rectangle, rounded corners, minimum width=0.8cm, minimum height=0.4cm,text centered,text width=0.8cm, draw=black, fill=blue!30]
\tikzstyle{C2B} = [rectangle, rounded corners, minimum width=2.3cm, minimum height=0.4cm,text centered,text width=2.3cm, draw=black, fill=blue!30]
\tikzstyle{C3B} = [rectangle, rounded corners, minimum width=1.8cm, minimum height=0.4cm,text centered,text width=1.8cm, draw=black, fill=blue!30]

\tikzstyle{C1Y} = [rectangle, rounded corners, minimum width=0.8cm, minimum height=0.4cm,text centered,text width=0.8cm, draw=black, fill=red!30]
\tikzstyle{C2Y} = [rectangle, rounded corners, minimum width=2.4cm, minimum height=0.4cm,text centered,text width=2.4cm, draw=black, fill=red!30]
\tikzstyle{C3Y} = [rectangle, rounded corners, minimum width=1.8cm, minimum height=0.4cm,text centered,text width=1.8cm, draw=black, fill=red!30]
\tikzstyle{C4Y} = [rectangle, rounded corners, minimum width=2.4cm, minimum height=0.4cm,text centered,text width=2.4cm, draw=black, fill=red!30]
\tikzstyle{arrow} = [thick,->,>=stealth]
\DeclareMathOperator*{\argmin}{arg\,min}

\newcommand{\bd}[1]{\boldsymbol{#1}}

\newcommand{\set}[1]{\mathcal{#1}}
\newcommand{\s}[1]{\mathcal{#1}}

\newtheorem{dfn}{Definition}

\title{An efficient $K$-means algorithm for Massive Data}
\author[1]{Marco Cap\'o\thanks{mcapo@bcamath.org}}
\author[1]{Aritz P\'erez\thanks{aperez@bcamath.org}}
\author[2]{Jos\'e A. Lozano\thanks{ja.lozano@ehu.eus}}
\affil[1]{Basque Center for Applied Mathematics, BCAM. Bilbao, Spain}
\affil[2]{ISG, University of the Basque Country UPV/EHU. San Sebastian, Spain}

\begin{document}

\maketitle
\begin{abstract}
Due to the progressive growth of the amount of data available in a wide variety of scientific fields, it has become more difficult to manipulate and analyze such
information. Even though datasets have grown in size, the {\it $K$-means} algorithm  remains as one of the most popular clustering methods, in spite of its dependency on 
the initial settings and high computational cost, especially in terms of distance computations. In this work, we propose an efficient approximation to the $K$-means problem intended for massive data. 
Our approach recursively partitions the entire dataset into a small number of subsets, each of which is characterized by its representative (center of mass) and weight
(cardinality), afterwards a weighted version of the $K$-means algorithm is applied over such local representation, which can drastically reduce the number of distances computed.
In addition to some theoretical properties, experimental results indicate that our method outperforms well-known approaches,
such as the {\it $K$-means++} and the {\it minibatch $K$-means}, in terms of the relation between number of distance computations and the quality of the approximation.

\end{abstract}






\section{Introduction}\label{sec:introduction}

The exponential increase of the data volumes that scientists, from different backgrounds, face on a daily basis implies the development of  simple yet scalable tools that eases the analysis and characterization of such information \cite{Jordan}. One of the most relevant analysis is data clustering. This process consists of grouping a given dataset into a predetermined amount of disjoint sets, called clusters. This is done in such a way that intra-cluster similarity is high and the inter-cluster similarity is low. Furthermore, clustering is a basic task of many areas, such as artificial intelligence, machine learning and pattern recognition \cite{Dubes,Jain}.

Even when there exists a wide variety of clustering methods, the $K$-means algorithm remains as one of the most popular \cite{Jain2}. In fact, it has been identified as one of the top $10$ algorithms in data mining \cite{Wu}.

\subsection{$K$-means}\label{sec:Kmeans}

Given a set of $n$ data points (instances) $D=\{\textbf{x}_1,\ldots,\textbf{x}_n\}$ in $\mathbb{R}^d$ and an integer $K$, the {\bf $K$-means problem} is to determine a set of $K$ centroids $C=\{\textbf{c}_1,\ldots,\textbf{c}_K\}$ in $\mathbb{R}^d$ so as to minimize the following {\bf error function}:

\begin{eqnarray}\label{eq:errorfunction}
E(C) = \sum\limits_{\textbf{x}\in D} \min\limits_{k=1,\ldots,K} \| \textbf{x}-\textbf{c}_k\|^2
\end{eqnarray}

This is a combinatorial optimization problem since it is equivalent to finding the partition
of the $n$ instances in $K$ groups whose associated set of centers of mass minimizes Eq.$1$.
In this case, the number of possible partitions is a Stirling number of the second kind \cite{Sami}.

Since finding the globally optimal partition is known to be {\bf NP-hard}, even for instances in the plane \cite{Aloise}, and exhaustive search methods are not useful in practice, iterative refinement based algorithms are commonly used to approximate the solution of the $K$-means and similar problems \cite{Kaufman, Lloyd, Park}. 
These algorithms iteratively relocate the data points between clusters until a locally optimal partition is attained. Among these methods the most popular is the { \bf $K$-means algorithm} \cite{Jain2, Lloyd}

The $K$-means algorithm has two stages: { \bf Initialization}, in which we set the starting set of centroids and, an iterative stage, called {\bf Lloyd's algorithm} \cite{Lloyd}. Lloyd's algorithm consists of two steps: A first step in which each instance is assigned to its closest centroid ({\bf assignment step}), then the set of centroids is updated ({\bf update step}). Finally, a stopping criterion
is verified. The most common criterion implies the computation of the error function (Eq.$1$): if the error does not decrease significantly, with respect to the previous iteration, the algorithm stops. The time required for the assignment step is  $O(nKd)$, while the update step of the set of centroids requires $O(nd)$ computations and
the stopping criterion based on the computation of the error function has a $O(nd)$ time complexity. Hence, the assignment step is the most computationally intensive due to 
the distance computations. For this reason, the main objective of our proposal consists of defining a variant of the $K$-means algorithm that controls the trade-off between
the number of distances computed and the quality of the obtained solution.

Conveniently, every step of the $K$-means algorithm can be easily parallelized \cite{Zhao}, which is a major key to meet the scalability of the problem \cite{Wu}.

\subsubsection{$K$-means Initialization}\label{sec:KI}

It is widely reported in the literature that the performance of the Lloyd's algorithm highly depends upon the initialization stage \cite{Lozano}: One might need several re-initializations before achieving a solution of acceptable quality. This is especially adverse when dealing with massive data applications since the number of distance computations is proportional to the number of instances, $n$. In addition, a poor initialization could lead to an exponential running time in the worst case scenario \cite{Vattani}. All these features are major downsides and show the importance of defining an appropriate initialization strategy.

There exist several approaches to initialize the $K$-means algorithm. One of the earliest, and most popular initialization strategies, was proposed by Forgy in $1965$ \cite{Forgy}. It consists of defining the initial centroids set as $K$ randomly selected instances from the dataset. The intuition behind this approach is that, by choosing the prototypes randomly, we are more likely to choose a point near an optimal cluster center, since such points tend to be where the highest density points are located \cite{Redmond}. The main disadvantage of this approach is that there is no guarantee that two, or more, of the selected seeds will not be near the center of the same cluster \cite{Redmond}.

Moreover, there also exist well known initialization procedures that are based on simple probabilistic seeding techniques. In particular, the {\it $K$-means++} method,
proposed by Arthur and Vassilvitskii in \cite{Arthur}, consists of randomly selecting only the first centroid from the dataset. Each subsequent initial centroid is 
chosen with a probability proportional to the distance with respect to the previously selected set of centroids. The key idea of this cluster initialization technique
is to preserve the diversity of seeds while being robust to outliers. The $K$-means++ algorithm leads to an $O(\log K)$ approximation of the optimal error after the 
initialization \cite{Arthur}. The drawback of this approach refers to its sequential nature, which hinders its parallelization, as well as to the fact that it 
requires $K$ scans of the entire 
dataset, therefore it has a complexity of $O(nKd)$.

\subsection{An alternative to Lloyd's algorithm}\label{sec:MBK} 

Apart from Lloyd's algorithm there exist several low computational cost alternatives that, by using statistical techniques, attempt to approximate
a suboptimal solution to the $K$-means problem without processing the information of every instance in the dataset \cite{Bradley, Davidson, Sculley}.
Among these algorithms we have the {\it minibatch $K$-means} proposed by Sculley in \cite{Sculley}. 
This algorithm seeks to reduce the computational cost by not using all the dataset at each iteration but small random batches of examples of a fixed size until convergence.
This strategy reduces the number of distance computations per iteration at the cost of lower cluster quality. Empirical results, in a range of large web based applications, 
corroborate that a substantial saving of computational time can be obtained at the expense of some loss of cluster quality \cite{Sculley}.

\subsection{Contribution}\label{sec:CTB} 

In this work, we propose an approximation algorithm for the $K$-means problem based on a recursive data partitioning process that reduces the number
of distance calculations and data scans, while generating competitive approximations. The algorithm considers a sequence of partitions of the dataset, in such a way that
the partition at iteration $i$ is thinner than the partition at iteration $i-1$. At each step, the mass center of each subset of the partition (set of representatives) is calculated and
a weighted Lloyd's algorithm is applied using the current set of representatives as the dataset. Among other benefits, this approach reduces the number of distance computations over the whole dataset, which is the most computationally demanding 
stage of the $K$-means algorithm. 

The rest of this article is organized as follows: In Section $2$, we describe the idea behind our algorithm and introduce notation that we
use, in Section $3$, to state some theoretical
guarantees of our approach. The proofs of such statements can be found in Appendix A. 
In Section $4$, we present a set of experiments in which we analyze the effect of different factors, such as the size of 
the dataset and the dimension of the instances over the performance of our algorithm. Additionally we compare these results with the ones 
obtained by the $K$-means++ and the minibatch $K$-means methods. Finally, in Section $5$, we define the next steps and possible improvements to our current
work.

\section{Recursive partition based $K$-means}\label{sec:contribution}

We propose a novel, iterative approximation strategy for the $K$-means problem that is based on a sequence of recursive partitions of the dataset, being
each partition thinner than the previous one. We call this approach {\bf recursive partition based $K$-means} ({\bf RP$K$M}). The idea behind the algorithm is to approximate the $K$-means problem for the full dataset by 
recursively applying a weighted version of the $K$-means algorithm over a growing, yet small, number of subsets of the dataset.

In the first step of the RP$K$M, the dataset is partitioned into a number of subsets each of which is characterized by a representative (center of mass) and its corresponding weight (cardinality).
Finally, a weighted version of  Lloyd's algorithm (see Section $2.2$ for further details) is applied over the set of representatives. From one iteration
to the next, a more refined partition is constructed and the process is repeated using the optimal set of centroids obtained at the previous iteration as initialization. 
This iterative procedure is
repeated until a certain stopping criterion is met.

In the next section, we describe in detail the recursive partition process and characterize some of its properties. The notation introduced in this section will be used later 
for a formal description of the RP$K$M algorithm.

\subsection{Recursive Partitions}\label{sec:Partition}

As previously mentioned, the recursive partition process is the first stage of the RP$K$M algorithm. This step consists of generating a thinner 
partition than the previous one at each iteration.

From now on we will use the following definition of partition of a dataset.

\begin{dfn}[{\bf Partition} of a dataset $D$]
$\mathcal{P}=\{S_1, \ldots, S_t  \}$ is a partition of the dataset $D$ if 
$\ \bigcup\limits_{i=1}^{t} S_i=D$ and if the subsets of $\mathcal{P}$ are (pairwise) disjoint and nonempty.
Moreover, given a subset $S \in \mathcal{P}$, we define its weight as its cardinality, $|S|$,
and its representative as its center of mass, $\overline{S}=\frac{\sum\limits_{{\bf x}\in {S}} \bf{x}}{|S|}$.
\end{dfn}

The partition of the dataset allows us to describe it with a reduced number of representatives, which ultimately implies the reduction of the number of distance computations with
respect to the Lloyd's algorithm for the full dataset. In the RP$K$M algorithm, we will use the set of representatives and weights, rather than the partition
itself.

\begin{dfn}[Partition {\bf thinner} than $\mathcal{P}$]
Given two partitions of the dataset $D$, $\mathcal{P}$ and $\mathcal{P}^{'}$, we say that $\mathcal{P}^{'}$ is a partition thinner than $\mathcal{P}$ ($\ \mathcal{P} \succ \mathcal{P}^{'}$) if,  
for all $S \in \mathcal{P}$, $S= \bigcup\limits_{R \in \mathcal{P}^{'}[S]}R$, where $\mathcal{P}^{'}[S]=\{R \in \mathcal{P}^{'}: R \subseteq S\}$.
\end{dfn}

In other words, $\mathcal{P}^{'}$ is a partition thinner than $\mathcal{P}$ if every subset of $\mathcal{P}$
can be written as the union of subsets of $\mathcal{P}^{'}$.

The partition process generates a {\bf sequence of thinner partitions}  $\ \mathcal{P}_1, \ldots, \mathcal{P}_m$, such that
$\ \mathcal{P}_{i-1} \succ \mathcal{P}_{i}$ for all $i\in \{2,\ldots,m\}$. Evidently, the number of representatives tends to increase as we generate a thinner partition.
In the extreme case $\mathcal{P}_m=\{\{{\bf x}\}:{\bf x} \in D\}$, however, in practice, in order to reduce the computational complexity of the RP$K$M, 
we control the number of representatives so that $|\mathcal{P}_m| \ll n$.

Note that the weight and the representative of $S \in \mathcal{P}_i$ can be easily computed from $\mathcal{P}_{i+1}[S]$ as follows:
$|S|=\sum\limits_{R\in \mathcal{P}_{i+1}[S]}|R|$, $\overline{S}=\frac{\sum\limits_{R\in \mathcal{P}_{i+1}[S]}|R|\cdot \overline{R}}{|S|}$. As we noted before,
we are interested in the computation of the set of representatives and weights, thus,
we will use $\mathcal{P}_{m}$ to generate the set of representatives and weights of the entire sequence of thinner partitions backward, from $\mathcal{P}_{m-1}$ to
$\mathcal{P}_{1}$. Hence, the construction of $\mathcal{P}_i$ has a $O(|\mathcal{P}_{i+1}| d)$ time cost 
for $i<m$. Moreover, if the assignment criterion of each instance of $D$ into its corresponding subset in $\mathcal{P}_{m}$ is of order $O(d)$, as it is
in the case of the grid based RP$K$M (see Section \ref{sec:Example}), then the construction of $\mathcal{P}_{m}$ is $O(nd)$ 
and, therefore, the cost of the entire partition process is $O(d(n+\sum_{i=2}^{m}|\mathcal{P}_i|))$.

\subsection{Weighted $K$-means problem}\label{sec:WKMP}
In this section, we introduce a generalization of the $K$-means problem defined over a set of weighted points, e.g. the set of representatives
and their respective weights associated to a partition. As a first step, we define a clustering for a partition.

\begin{dfn}[Clustering for a partition $\mathcal{P}$]
We say that a partition of the dataset $D$, $\mathcal{G}$, is a clustering of the dataset for a partition $\mathcal{P}$, when $|\mathcal{G}|=K$ and $\ \mathcal{G} \succ \mathcal{P}$.
\end{dfn}

In other words, a cluster for a partition is a set of $K$ subsets of points of $D$, such that all the points of any $S\in \mathcal{P}$ are
assigned to the same cluster.

We call $\set{G}=\{G_1,..,G_K\}$ a {\bf clustering induced by a set of centroids} $C=\{\textbf{c}_1,...,\textbf{c}_k\}$, when $G_{k}= \bigcup\limits_{S \in M_{k}}S$ for $k=1,...,K$, where $M_{k}=\{S \in \mathcal{P}: k=\argmin\limits_{j=1,\ldots,K} \| \overline{S}- \textbf{c}_{j} \|^2 \}$. In other words, a clustering induced by a set of centroids is a partition of the dataset in which all the data points that have the same closest centroid from $C$ are grouped in the same cluster. We denote that the clustering $\set{G}$ is induced by a set of centroid $C$ by $\set{G}\leftarrow C$. Similarly, we call $C=\{\textbf{c}_1,...,\textbf{c}_k\}$ a {\bf centroids set induced by a clustering} $\set{G}=\{G_1,...,G_K\}$, when $c_i= \overline{G}_i$ for $i=1,...,K$.
In other words, the set of centroids induced by a clustering $\set{G}$ is the set of centers of mass associated to each cluster in $\set{G}$.
We denote that the set of centroids $C$ is induced by a clustering $\s{G}$ by $C \leftarrow \set{G}$.

Given a partition of the dataset $D$, $\mathcal{P}$, the {\bf weighted $K$-means problem} seeks to determine a set of $k$ centroids $C=\{\textbf{c}_1,\ldots,\textbf{c}_K\}$ in $\mathbb{R}^d$, so as to  minimize the {\bf centroid error} associated to a partition $\mathcal{P}$, which is defined as follows:
\begin{eqnarray}
E_{\mathcal{P}}(C)&=&\sum\limits_{S \in \mathcal{P}} |S| \min\limits_{k=1,\ldots,K} \| \overline{S}- \textbf{c}_k \|^2\nonumber\\
&=&\sum_{k=1}^K \sum_{S \in \s{P}: S \subseteq G_k} |S|\cdot||\overline{S}-\bd{c}_k||^2\label{eq:centroidError}
\end{eqnarray}
where the clustering $\set{G}$ is induced by the set of centroids $C$. This error measures the weighted error between the representative of each subset with respect to its closest centroid.

\begin{algorithm}[H]
    \caption{{\bf Weighted Lloyd (WL)}}\label{alg:wlloyd}

   {\bf Input:} Set of representatives $\{\overline{S}\}_{S \in \mathcal{P}}$ and weights $\{|S|\}_{S \in \mathcal{P}}$, for \\ 
   $\ \ \ \ \ \ \ \ \ \ $ the partition $\mathcal{P}$. Number of clusters $K$ and initial set of \\
   $\ \ \ \ \ \ \ \ \ \ $ centroids $C_{0}$.

   {\bf Output:} Set of centroids $C_{r}$ and corresponding clustering pattern $\mathcal{G}_{r}$.

   {\bf Step 0} ({\bf Initial Assignment}): \\
   $\ \ \ \ \ \ \ \ \mathcal{G}_{0}\longleftarrow C_{0}$; \ $r=0$.\\
   \While{not StoppingCriterion}{
   $r=r+1$.\\
   {\bf Step 1} ({\bf Update Step}):$\ \ \ \ \ \ \ \ \ \ \ \ \ \ \ \ \ \ \ \ \ \ \ \ \ \ \ \ \ \ \ \ \ \ $ {\bf Clustering error} \\
   $\ \ \ \ \ \ \ \ \ \ $ $C_{r}\longleftarrow \mathcal{G}_{r-1} $ $\ \ \ \ \ \ \ \ \ \ \ \ \ \ \ \ \ \ \ \ \ \ \ \ \ \ \ \ \ \ \ \ \ \ \ \ \ \ $ $ E_{\mathcal{P}}(\mathcal{G}_{r-1})$  (Eq. \ref{eq:clusteringError})\\
{\bf Step 2}  ({\bf Assignment Step}): $\ \ \ \ \ \ \ \ \ \ \ \ \ \ \ \ \ \ \ \ \ \ \ \ $ {\bf Centroid error} \\
   $\ \ \ \ \ \ \ \ \ \ $  $\mathcal{G}_{r}\longleftarrow C_{r}$ $\ \ \ \ \ \ \ \ \ \ \ \ \ \ \ \ \ \ \ \ \ \ \ \ \ \ \ \ \ \ \ \ \ \ \ \ \ \ \ \ \ \ \ $ $ E_{\mathcal{P}}(C_{r})$ (Eq. \ref{eq:centroidError})
}
Return $C_{r}$ and $\mathcal{G}_{r}$.
\end{algorithm}

In order to approximate the solution of the weighted $K$-means problem, we use a generalization of Lloyd's algorithm called the {\bf weighted Lloyd's algorithm} ({\bf WL}, see Algorithm \ref{alg:wlloyd}). In the assignment stage of WL (Step $0$ and Step $2$), the clustering $\mathcal{G}_{r}$ is induced by the set of centroids $C_r$. Furthermore, in the update step (Step $1$), the set of centroids $C_r$ is induced by the clustering $\set{G}_{r-1}$. Similarly to Lloyd's algorithm, an execution of WL with $l$ iterations produces a sequence of sets of centroids and clusterings that can be represented as follows: 
$$C_0\rightarrow \set{G}_0\rightarrow C_1\rightarrow \set{G}_1 \rightarrow...\rightarrow C_{l-1}\rightarrow \set{G}_{l-1} \rightarrow C_l \rightarrow \set{G}_l$$
where $C_0$ is the set of centroids used for initialization and $C_l$ is the returned set of centroids.

The assignment step requires $O(|\mathcal{P}|kd)$ computations, since we just need to compute the distance between the set of centroids and the set of representatives, while for the update step of the set of centroids and the computation of its error (centroid error) $O(|\mathcal{P}|d)$ computations are needed. Remember that the most common stopping criterion of the $K$-means algorithm consists of verifying that the difference of the set of centroids error, in two consecutive iterations, is smaller than a certain threshold. Moreover, observe that the set of weights is only used when updating the set of centroids. Since the number of representatives usually satisfies $|\mathcal{P}| \ll n$, when dealing with massive data problems, we can have a relevant reduction in the complexity with respect to the $K$-means algorithm for the full dataset.

\subsection{RP$K$M Algorithm}\label{sec:Algorithm}

In this section, we formally present the RP$K$M algorithm. This algorithm mainly consists of constructing a sequence of thinner partitions $\ \mathcal{P}_1, \ldots, \mathcal{P}_m$ and then applying WL over the set of representatives of each partition in the sequence. From one iteration to the next, the preceding found solution is used as initialization. As we will show later, this initialization assignment allows us to reduce the maximum number of WL iterations at every RP$K$M run. The pseudo-code of the RP$K$M algorithm can be seen in Algorithm $1$.

\vspace{0.2cm}

\begin{algorithm}[H]
    \caption{{\bf RP$K$M Algorithm}}

   {\bf Input:} Dataset $D$, number of clusters $K$, maximum number of \\ 
   $\ \ \ \ \ \ \ \ \ \ $ iterations $m$.

   {\bf Output:} Set of centroids approximation $C_i$.
   
   {\bf Step 1} Compute the set of weights and representatives of the \\
   $\ \ \ \ \ \ \ \ \ \ $ sequence of thinner partitions,
   $\ \mathcal{P}_1, \ldots, \mathcal{P}_m$, backwards. \\
   $\ \ \ \ \ \ \ \ \ \ $ Set $i=1$.

   \While{not Stopping Criterion}{

{\bf Step 2}  Update the centroid's set approximation, $C_{i}=\{{\bf c}_j^i\}_{j=1}^K$:\\

$\ \ \ \ \ \ \ \ \ \ C_i$= {\it WL}$(\{\overline{S}\}_{S \in \mathcal{P}_i},\{|S|\}_{S \in \mathcal{P}_i},K,C_{i-1})$ \\
$i=i+1$

}
Return $C_{i}$
\end{algorithm}

\vspace{0.2cm}

In the first step of the RP$K$M algorithm, we obtain backwards (see Section $2.1$) the set of representatives and weights associated to the sequence of thinner partitions $\ \mathcal{P}_1, \ldots, \mathcal{P}_m$. Observe that we are assuming, without loss of generality, that $|\mathcal{P}_{1}|>K$. In Step $2$, we update the centroids approximation by applying WL using the representatives and weights set determined at the previous step, we take as initialization the approximation for the previous iteration, $C_{i-1}$. In the first RP$K$M iteration, we set $C_{i-1}$ as $K$ random representatives of $\{\overline{S}\}_{S \in \mathcal{P}_i}$ (Forgy's type initialization). The algorithm iterates until $i=m$ or until a stopping criterion is met. We recommend the computation of a centroid's set displacement measure as stopping criterion: $\delta({C_{i-1},C_{i}})= \max\limits_{j=1,...,K} \|{\bf c}_j^i-{\bf c}_j^{i-1} \|^2$. If this value is smaller than a certain threshold, the algorithm stops, since the approximation did not improve significantly after the last RP$K$M iteration.

In relation to the complexity of Algorithm $2$, we know, from Section $2.1$, that Step $1$ has an $O(d(n+\sum_{i=2}^{m}|\mathcal{P}_i|))$ time cost. Moreover, at the $i$-th RP$K$M iteration, the time required for WL (Step $2$) is $O(|\mathcal{P}_i| K d)$. Finally, the recommended stopping criterion just performs $O(Kd)$ computations. Hence, the overall complexity of the RP$K$M algorithm, in the worst case, is $O(\max \{d(n+\sum_{i=2}^{m}|\mathcal{P}_i|),|\mathcal{P}_m| K d\})$.

\subsection{RP$K$M implementation based on grid partitions}\label{sec:Example}

Later on, we will verify that the theoretical advantages of the RP$K$M algorithm hold independently of the geometry that we use to generate the partition. Nonetheless, one way to guarantee the generation of a sequence of thinner partitions of the dataset consists of partitioning the space in a recursive manner. To do so, one possibility is to use a generalization of the {\it quadtrees} for higher dimensions \cite{Finkel}. The quadtree data structure has been used in several areas such as dimension reduction problems, spatial indexing, storing sparse data, computer graphics: computational fluid dynamics, etc \cite{Gandhi}.

The $d$-dimensional generalization of a quadtree is a tree data structure that generates partitions of the space into $d$-dimensional hypercubes and, subsequently, of the dataset in subsets in the following way: each internal node of the tree  is exactly divided  in $2^{d}$ children, i.e., each subset of the $i$-th partition is divided into, at most, $2^d$ sets of the $(i+1)$-th partition (see Fig.$1$). This property allows us to generate, in a simple manner, a sequence of thinner partitions at each iteration satisfying $|\mathcal{P}_i| \leq \min\{n,2^{id}\}$. 

\begin{figure}[H]
 \begin{center}
        \includegraphics[height=0.38\textheight,width=0.85\textwidth]{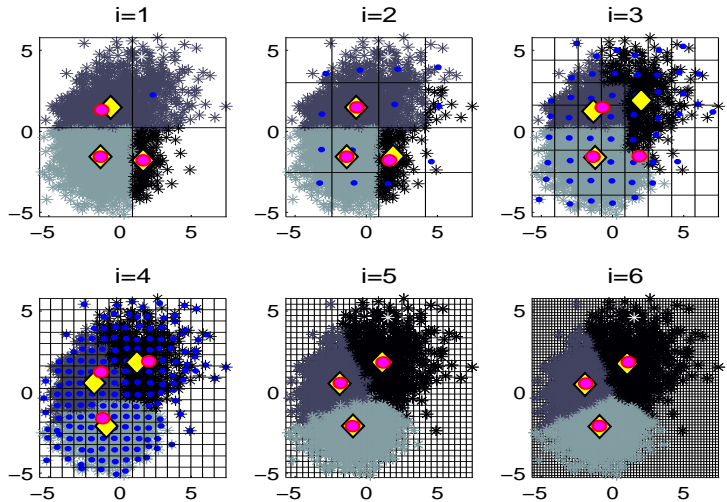}
     \vspace{-0.7cm}
        \caption{Best clustering obtained at each RP$K$M iteration}
        \label{figu1}
         \end{center}
        \end{figure}

In the following example, we consider a set of $10000$ points generated from a mixture of three $2D$ Gaussians. We compute, as a reference, the solution for $K=3$ using the $K$-means++ method. After ten runs, we obtained, on average, an error of $11393.45$ with a standard deviation of $4.69$. The number of distance computations was, on average, $642000$. In Fig.\ref{figu1}, we show the evolution of the RP$K$M algorithm, for $m=6$, the red circles represent the initial set of centroids, the yellow diamonds the final set of centroids and the blue points the set of representatives for each iteration.

\begin{table}[H]
\addtolength{\tabcolsep}{1.6pt}
 \begin{minipage}{1\linewidth}
  \begin{center}
\begin{tabular}{|c|c|c|c|}
\hline
{\bf $i$} & {\bf Dis Com} & {$|\mathcal{P}_i|$} &  {$E(C_i)$}\\ \hline
$1$ & $24$ & $4$ & $14050.06$\\
$2$ & $114$ & $15$ & $14024.38$\\
$3$ & $1545$ & $53$ & $12350.41$\\
\hline
\end{tabular} \quad
\begin{tabular}{|c|c|c|c|}
\hline
{\bf $i$} & {\bf Dis Com} & {$|\mathcal{P}_i|$} &  {$E(C_i)$}\\ \hline
$4$ & $5697$ & $173$ & $11424.24$\\
$5$ & $10449$ & $528$ & $11408.40$\\
$6$ & $26781$ & $1361$ & $11389.54$\\
\hline
\end{tabular}
\end{center}
\vspace{-0.5cm}
\caption{RP$K$M iteration results}\label{table:results}
\end{minipage}
\end{table}

From Table $1$, we can observe that, even at the fourth grid based RP$K$M iteration, which in this case implies $173$ representatives ($1.73 \%$ of the dataset),
we have a fairly good approximation of the average best solution found by the $K$-means++ algorithm for the entire $10000$ points. On average, the RP$K$M algorithm
computed $0.887\%$ and $4.17\%$ of the total number of distance computations of the $K$-means++ algorithm, at the fourth and final iteration respectively.

As we consider higher iterations of the RP$K$M,  the associated cost function converges to the best solution obtained by the the $K$-means++.
The intuition behind this method is to transform a random initial set of centroids 
into a competitive approximation by using small groups of representatives, instead of the entire dataset. Next, we consider higher values of $i$ to refine
such an approximation.

\section{Theoretical analysis of the RP$K$M algorithm}\label{sec:properties}
In this section, we perform a theoretical analysis of RP$K$M. In Section \ref{sec:evolution}, we analyze the evolution of the clustering error at different steps of RP$K$M. Then, in Section \ref{sec:bounds}, we investigate the repetitions of the clusterings obtained during the execution of RP$K$M and we bound the maximum number of WL iterations for different steps of RP$K$M.

Before starting with the theoretical results, we summarize an execution of RP$K$M with $m$ steps given in terms of sequences of centroids and clusterings:
\begin{eqnarray}\label{eq:summaryRPKM}
\set{P}_1:&& \ C_0^1 \rightarrow \set{G}_0^1 \rightarrow C_1^1 \rightarrow \set{G}_1^1 \rightarrow ... \rightarrow \set{G}_{{l_1}-1}^1 \rightarrow C_{l_1}^1 \rightarrow \set{G}_{l_1}^1 \nonumber\\
\set{P}_2:&& \ C_0^2 \rightarrow \set{G}_0^2 \rightarrow C_1^2 \rightarrow \set{G}_1^2 \rightarrow ... \rightarrow \set{G}_{{l_2}-1}^2 \rightarrow C_{l_2}^2 \rightarrow \set{G}_{l_2}^2 \nonumber\\
...\nonumber\\
\set{P}_i:&& \ C_0^i \rightarrow \set{G}_0^i \rightarrow C_1^i \rightarrow \set{G}_1^i \rightarrow ... \rightarrow \set{G}_{{l_i}-1}^i \rightarrow C_{l_i}^i \rightarrow \set{G}_{l_i}^i \nonumber\\
...\nonumber\\
\set{P}_m:&& \ C_0^m \rightarrow \set{G}_0^m \rightarrow C_1^m \rightarrow \set{G}_1^m \rightarrow ... \rightarrow \set{G}_{{l_m}-1}^m \rightarrow C_{l_m}^m \rightarrow \set{G}_{l_m}^m
\end{eqnarray}
where $l_i$ corresponds to the number of iterations of WL at step $i$ of RP$K$M, the set of centroids $C_{r+1}^i$ is induced by the clustering $\set{G}_{r}^i$, and $\set{G}_r^i$ is induced by $C_r^i$ for $r=1,...,l_i-1$ and $i=1,..,m$. Each line corresponds to an execution of WL for a given partition $\set{P}_i$ for $i=1,...,m$. It should be noted that, in step $i$ of RP$K$M, the set of centroids $C_0^i$ corresponds to the set of centroids obtained at the end of its previous step, that is $C_0^i=C_{l_i-1}^{i-1}$ for $i=1,...,m$. However, the clustering induced by $C_0^i=C_{l_i-1}^{i-1}$ for partition $\set{P}_i$ does not have to correspond to the clustering induced for the previous partition $\set{P}_{i-1}$. This fact is one of the main difficulties in guaranteeing a monotone decrement of the error function (see Eq.\ref{eq:errorfunction}) during an execution of RP$K$M.

In order to analyze the behavior of RP$K$M, we define the {\bf clustering error} associated to a partition $\set{P}$ as follows:
\begin{equation}
E_{\mathcal{P}}(\set{G})= \sum_{k=1}^K \sum_{S \in \s{P}: S \subseteq G_k} |S|\cdot||\overline{S}-\bd{c}_k||^2\label{eq:clusteringError}
\end{equation}
where the set of centroids $C$ is induced by the clustering $\set{G}$. This function measures the weighted error between each representative of a partition $\set{P}$ and the center of mass of its corresponding cluster. Note that the only difference between the centroid error and clustering error is that, the centroid error is given in terms of a set of centroids and its induced clustering, while the clustering error is given by a clustering and its induced set of centroids. The importance of the clustering error is that it represents an intermediate value between the centroid errors obtained at two consecutive iterations of the algorithm, that is 
\begin{equation}\label{eq:inequality}
E_{\set{P}_i}(C_r^i) \geq E_{\set{P}_i}(\set{G}_r^i) \geq E_{\set{P}_i}(C_{r+1}^i) 
\end{equation}
for $r=0,...,l_i-1$ (see Eq.\ref{eq:summaryRPKM}). In the following subsections we will analyze the relation between the centroid error for different partitions of the data based on the inequality provided in Eq.\ref{eq:inequality}.

\subsection{Evolution of the centroids error}\label{sec:evolution}
In this section we analyze the evolution of the centroid error for RP$K$M. The obtained results will be the basis for bounding the number of iterations of WL at each step of the RP$K$M. The next result will be used in order to analyze the relation between the clustering error given two partitions of the dataset (one thinner than the other).
\begin{restatable}{lm}{primelemma}\label{lemma:constant}
Given a set of points $D$ in $\mathbb{R}^d$ and a partition of it, $\mathcal{P}$. Then the function 
$f(\textbf{c})= |D|\cdot\| \overline{D}- \textbf{c} \|^2 - \sum_{R \in \mathcal{P}} |R|\cdot\| \overline{R}- \textbf{c} \|^2 $ is constant.
\end{restatable}
This result implies that the difference of the set of representatives with respect to a centroid, for two partitions of the dataset (one thinner than the other), is constant. The fact that such a difference is constant allows us to state, in the following lemma, the invariance of the clustering error for two different partitions of the dataset. Observe that Lemma \ref{lemma:constant} allows us to change the clustering, for both partitions, without changing the difference of the error associated to them.

\begin{figure}[t]
\begin{center}
    \includegraphics[height=0.6\textwidth]{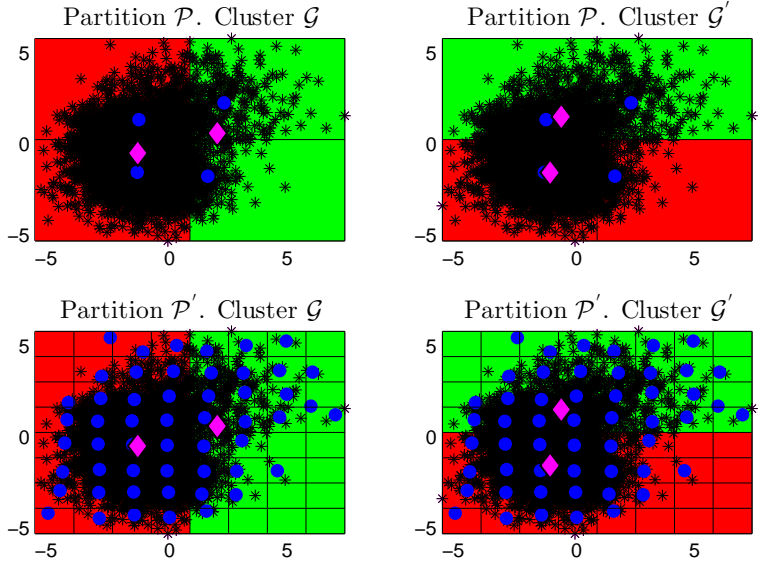}
    \caption{Illustration of Lemma \ref{lemma:invariance}, for two clusterings $\mathcal{G}$ and $\mathcal{G}^{'}$ defined on a partition $\mathcal{P}$}
    \label{esPARG}
\end{center}
\end{figure}

\begin{restatable}{lm}{seclemma}\label{lemma:invariance} [Invariance of the clustering error difference]
Let $\set{P}$ and $\set{P}'$ be two partitions of the dataset $D$, with $\ \mathcal{P} \succ \mathcal{P}^{'}$, and let $\s{G}$ and $\s{G}'$ be two clusterings of $\mathcal{P}$. Then, the difference between both clustering errors is constant with respect to the partitions $\s{P}$ and $\s{P}'$:
\begin{eqnarray*}
E_{\mathcal{P}}(\mathcal{G})-E_{\mathcal{P}}(\mathcal{G}^{'})&=&E_{\mathcal{P}^{'}}(\mathcal{G})-E_{\mathcal{P}^{'}}(\mathcal{G}^{'})
\end{eqnarray*}
\end{restatable}
In other words, the difference between two clustering errors is independent of the partition. For example, in Fig.\ref{esPARG}, clustering $\mathcal{G}$ restricts the subsets with center of mass to the left (right) of the middle point of the bounding box to belong to the same cluster. The pink diamonds represent the centers of mass of each group of $\mathcal{G}$, evidently such centers of mass are invariant with respect to the partition that we use to represent $\mathcal{G}$. Furthermore, Lemma \ref{lemma:invariance} states that the difference of the clustering error difference between the upper figures is equivalent to the difference of the error associated to the lower figures in Fig.\ref{esPARG}. 

In general, we can not guarantee a monotone descent of the error function given in Eq.\ref{eq:errorfunction}, which corresponds to the centroid error for the thinnest partition, i.e., $\s{D}=\{\{x\}: x \in D\}$. However, in the next result, under mild conditions related to the difference of the centroid error, we prove a monotone descent of the clustering error for two partitions, one thinner than the other. In consequence, if the conditions stated for the difference of the clustering error hold for all the steps of RP$K$M, a monotone descent of the error function in Eq.\ref{eq:errorfunction} is guaranteed.
\begin{restatable}{cor}{sectheo}\label{thm:monotonedescend}[Condition for a monotone descent of the centroid error] 
Let $C_i$ and $C_{i-1}$ represent the set of centroids obtained at the $i$-th and ($i-1$)-th RP$K$M step, respectively. Then $E(C_{i}) \leq E(C_{i-1})$, if and only if $E(\mathcal{G}_{l_{i-1}-1}^{i-1})- E(C_{i-1}) \leq \xi_{i} + (E(\mathcal{G}_{l_{i}-1}^{i})- E(C_{i}))$, where $\xi_i=E_{\mathcal{P}_{i}}(\mathcal{G}_{l_{i-1}-1}^{i-1})-E_{\mathcal{P}_{i}}(\mathcal{G}_{l_{i}-1}^{i})$
\end{restatable}

That is, if after the assignment step for both sets of centroids, $C_i$ and $C_{i-1}$, with respect to the full dataset, the condition in Theorem \ref{thm:monotonedescend} is satisfied at every RP$K$M iteration, then we can guarantee the monotone descent of the error over the entire dataset. In particular, if there are no reassignments for any of the two cases, with respect to their associated cluster membership, we can guarantee the monotone descent of the overall error. Clearly, as the difference of the local error of the initial and final cluster at the $i$-th RP$K$M step is larger, then it is more likely to satisfy such a condition.

Even when the monotone descent over the entire dataset of the RP$K$M approximation, at every step, is not proved in general, we will see in the experiments summarized in Section \ref{sec:experimental} that, for real and artificial datasets, this property has been observed.

\subsection{Bounding the iterations of the weighted Lloyd's algorithm}\label{sec:bounds}
In this section, using the properties of the clustering error (Lemma \ref{lemma:invariance}), we can analyze the construction of the set of clusterings at different RP$K$M steps, for example we can verify the implications of repeating a clustering from a previous RP$K$M step.

In Lemma \ref{lemma:norepetitions}, we state that the unique clustering that can be repeated in a step of RP$K$M, is the previous clustering of the sequence of clusterings generated by the RP$K$M. If the repeated clustering is obtained at the first iteration of WL, then the previous clustering corresponds to the one obtained at the last iteration of WL (at the previous step of RP$K$M). On the other hand, if the repeated clustering is not obtained at the first iteration of WL, then the previous clustering corresponds to the one obtained at the previous iteration of WL (in the same RP$K$M step).
\begin{restatable}{lm}{terlemma}\label{lemma:norepetitions}
At the $i$-th step of the RP$K$M, if $\mathcal{G}_{r}^{i}=\mathcal{G}_{s}^{j} $, with $j < i$, for some $r \in \{1,\ldots, l_{i}-1\}$ and $s \in \{1,\ldots, l_{j}-1\}$, then $l_{j+1}=\ldots=l_i = 1$. Moreover, in that case, $s=l_{j}-1$. 
\end{restatable}

In the following theorem, we use Lemma \ref{lemma:norepetitions} to bound the number of WL iterations at each RP$K$M step. Lemma \ref{lemma:norepetitions} indicates that the only cluster that can be repeated, from a previous RP$K$M step, is the last one generated by the corresponding WL execution. Therefore, we can eliminate, from the total number of possible clusterings, the ones that were generated at previous RP$K$M iterations (except the last one). In particular, if we have more than one Lloyd iteration at a certain RP$K$M step, then we automatically discard every single cluster that was generated at a previous RP$K$M step. 

\begin{restatable}{thm}{primtheo}\label{theo:UpperBound}[Upper bound to the number of local WL iterations] 
An upper bound to the number of Lloyd iterations at the $i$-th RP$K$M step is given by \ $l_i \leq {|\mathcal{P}_{i}| \brace K}-\sum\limits_{j=1}^{i-1} (l_j - 1)$, where ${|\mathcal{P}_{i}| \brace K}$ is a Stirling number of the second kind. 
\end{restatable}

Following the same reasoning as in Theorem \ref{theo:UpperBound}, observe that, if, at the $(i-1)$-th RP$K$M step, WL converges to the associated global optima, then $l_i \leq {|\mathcal{P}_{i}| \brace K}-{|\mathcal{P}_{i-1}| \brace K}+1$. Moreover, observe that all the clusters with an error greater than $E_{\mathcal{P}_i}(\mathcal{G}_{l_{i-1}-1}^{i-1})$ will not be generated in the current or at any further RP$K$M iteration, however the amount of clusterings satisfying this condition can not be counted at the moment, one hypothesis is that the number of such clusterings is of order $O({|\mathcal{P}_{i}| \brace K})$.

For this reason, selecting the local initialization of WL in this manner may help reducing the number of Lloyd's iterations, while discarding, at each step, all the generated clusters (except one) and probably others of similar form. Not only that, but the discarding of such clusters occurs while analyzing a small number of representatives with regard to the full dataset, which implies, as we will see in the experimental section, a drastic reduction in the number of distance computations.

\section{Experimental section}\label{sec:experimental}

In this section, we perform a set of experiments so as to analyze the relation between the number of distance computations and the quality of the approximation for the grid based RP$K$M algorithm proposed in Section $2$. In addition, we analyze the effect on the algorithm performance of varying the different parameters of the clustering problem: size of the dataset, $n$, dimension of the instances, $d$, and number of clusters, $K$. For the purposes of the experimental analysis, we compare the performance of the grid based RP$K$M algorithm against the {\bf $K$-means++} ({\bf $K$M++}) and the {\bf minibatch $K$-means} ({\bf MB}) on artificial and real datasets.

The grid based RP$K$M was implemented in Python, while we used the $K$M++ and MB implementations that are available in the open source machine learning library {\it scikit-learn} of Python. As stopping criterion for the RP$K$M, we just set a maximum number of iterations, $m$, since we want to analyze the behavior of the error function, at each step, as the number of representatives approaches the number of instances. For the analyzed datasets, we observe that, with $m\leq 6$, this occurs. Evidently, as we increase the dimension, this property will be seen inmediately, since the number of representatives increases exponentially with respect to this parameter.

In this section, we refer to the result obtained after the $m$-th step of the grid based RP$K$M by {\bf RP$K$M $m$}, and to the solution obtained using MB with a batch size $b \in \{100,500,1000\}$ by {\bf MB $b$}. In equivalent experimentations similar batch sizes were used \cite{Sculley}.

\subsection{Artificial datasets results}
The artificial datasets are generated as a $d$-dimensional mixture of $K$ Gaussians. In particular, we set $K\in\{3,9\}$, $d\in\{2,4,8\}$ and $n\in\{1000,10000,$ $100000,1000000\}$. For each setting,  we generate $50$ replicates of the dataset. Additionally, we consider a component overlapping lower than $5\%$.

\subsubsection{Distance computations}
In this section, we compare the behavior of RP$K$M, $K$M++ and MB in terms of the computed distances. As we commented in Section \ref{sec:Kmeans} and Section \ref{sec:WKMP}, the most time consuming phase of the Lloyd's algorithm, and its weighted version, refers to the computation of distances. Especially at the initial steps, RP$K$M considers a number of representatives which is a small fraction of the size of the dataset. Thus, we would expect a greater reduction in the number of distance computations, with respect to the other methods, as we consider larger datasets. 

\begin{figure}[H]
 \begin{center}
        \includegraphics[height=0.55\textheight,width=\textwidth]{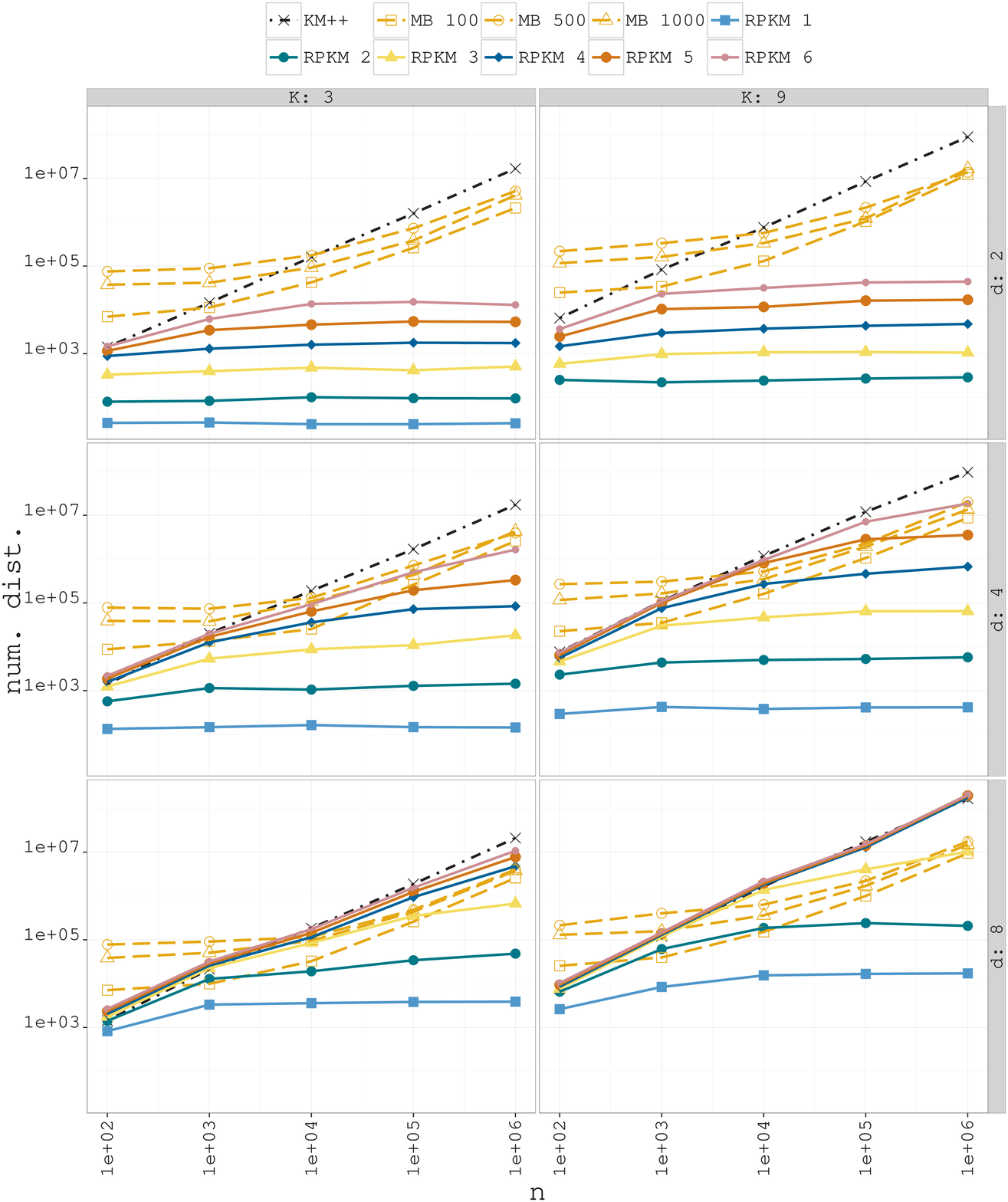}
        \caption{This figure shows the number of distance computations with respect to the size of the dataset ($n$), for different numbers of dimensions ($d$) and numbers of clusters ($K$)}
        \label{figu3}
         \end{center}
        \end{figure}

In Fig.\ref{figu3}, we present the relation between the number of distance computations and the dataset size for the different settings. 
\vspace{-0.6cm}

At first glance, we observe that  RP$K$M, in general, executes a much smaller number of distance computations than both $K$M++ and MB. Such a relation seems to change for the latter steps of RP$K$M when we consider larger dimensions. However, in that case, $K$M++ still requires a similar order of computations in comparison to the latter steps of the RP$K$M. Analogously, MB, for the different batches, is not able to match the number of distance computations 
of RP$K$M at its first steps, for any of the analyzed settings. In addition, we observe that, for some RP$K$M steps, the number of distance computations does not increase as we consider a higher number of instances, as happens with the other algorithms.  

\begin{figure}[H]
 \begin{center}
        \includegraphics[height=0.55\textheight,width=\textwidth]{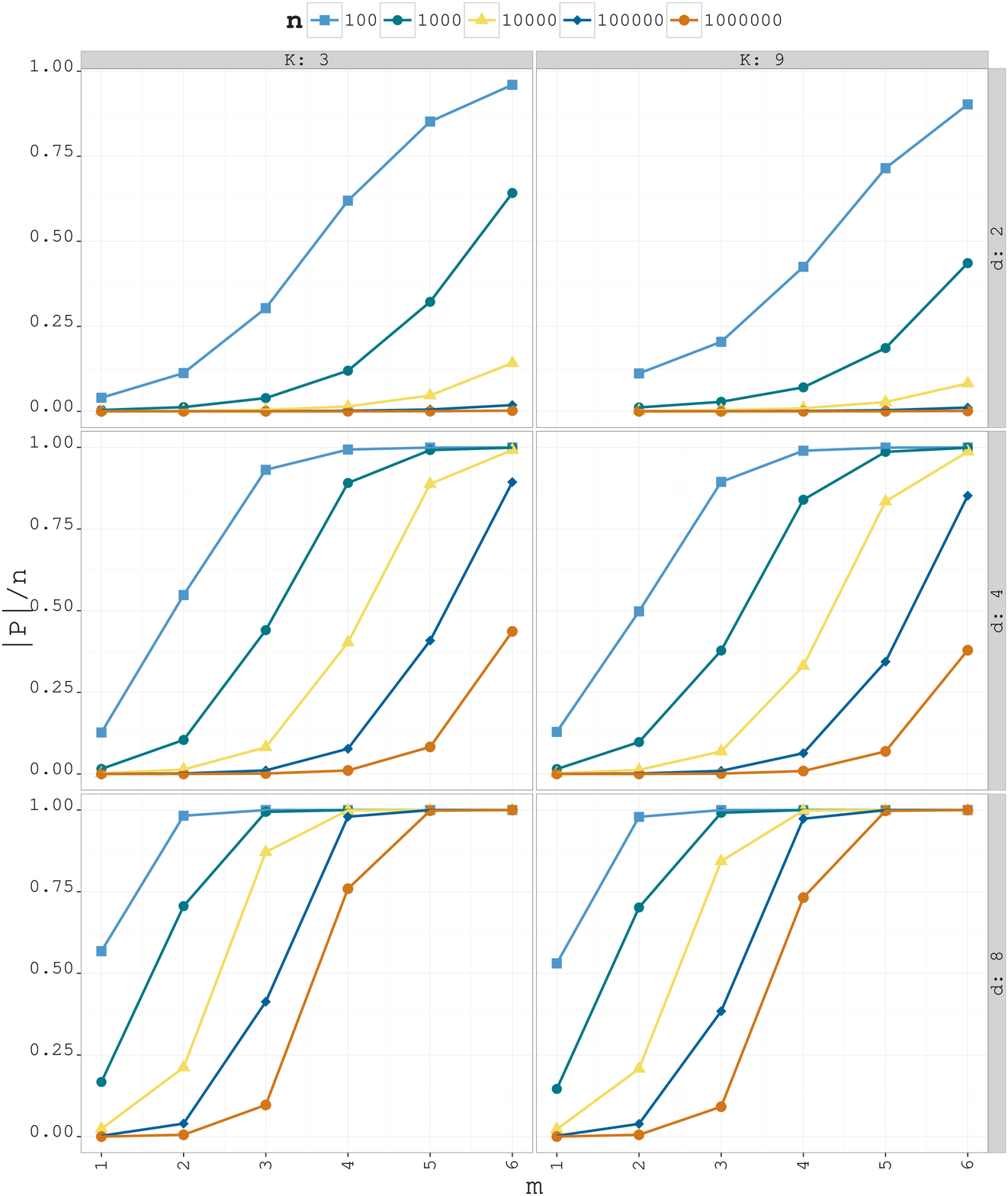}
     \vspace{-0.2cm}
        \caption{This figure shows the ratio between the size of the partition $P_m$ and the size of the data set $n$, for $m=1,...,6$. The size of the partition $\set{P}_m$ corresponds to the number of representatives used by RP$K$M at its $m$-th step.}
        \label{figu4}
         \end{center}
        \end{figure}

In particular, we notice that the number of distance computations, at the first steps of the RP$K$M, i.e., RP$K$M $1$ and RP$K$M $2$, does not necessarily increase with respect to the dataset size. This is plausible since, in this case, the number of representatives is of the same order, independent of the number of instances (see Fig.\ref{figu4}). Evidently, as we consider thinner partitions ($m \geq 3$), the number of representatives will increase and so will the number of distance computations.

\vspace{-0.6cm}
Since, at the earlier stages of the RP$K$M, the number of representatives does not necessarily increase with respect to the dataset size, then the ratio between the number of distance computations of RP$K$M and $K$M++ (or MB) decreases with respect to the size of the dataset. In particular, for the larger number of instances, the number of distances computed by RP$K$M with respect to $K$M++ is $3$ orders of magnitude lower for $K=3$ and $d=8$ and $6$ orders of magnitude lower for $K=3$ and $d=2$. In comparison with MB, the number of distances computed by RP$K$M is $2$ orders of magnitude lower for $K=3$ and $d=8$ and $5$ orders of magnitude lower for $K=3$ and $d=2$. As we can see, the dimensionality of the problem, $d$, has a great impact on the number of distances computed by RP$K$M as $m$ increases. The reason is that the number of representatives used by RP$K$M can increase exponentially with respect to both $m$ and $d$.
In addition, we can see that the number of distance computations, for all the algorithms, as expected, increases linearly with the number of clusters $K$.

\subsubsection{Quality of the approximation}
In the previous section, we observed that RP$K$M entails a drastic reduction in the amount of distance computations with respect to the other approaches (especially when we consider large dataset sizes). However, in this section, we would like to analyze the quality of the approximations obtained by means of RP$K$M.

\begin{figure}[H]
 \begin{center}
        \includegraphics[height=0.60\textheight,width=\textwidth]{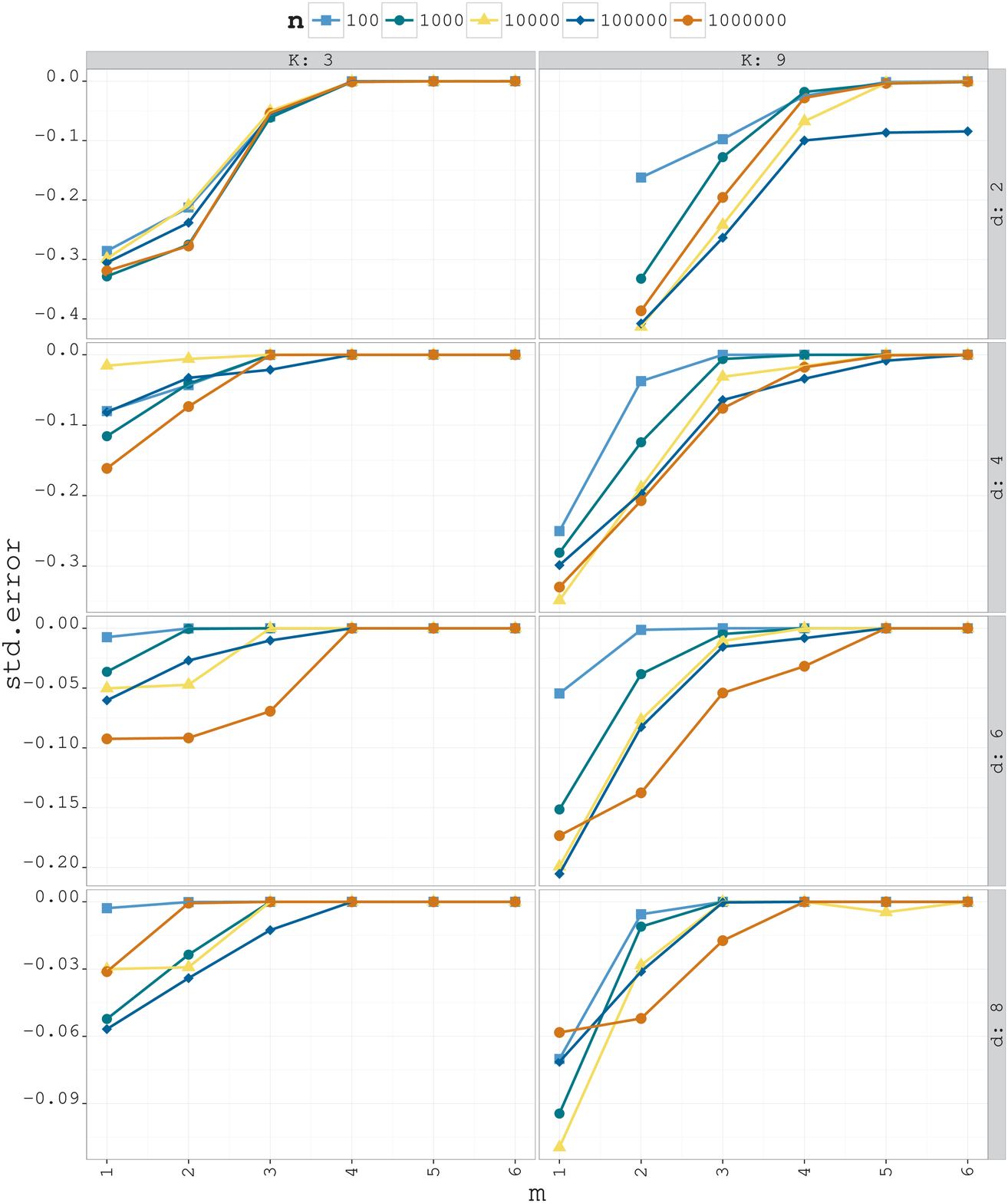}
        \caption{Quality of the approximation (std.error) with respect to the RP$K$M step}
        \label{figu5}
         \end{center}
        \end{figure}

In Fig.\ref{figu5}, we show the evolution of the standarized error (std.error) for the full dataset for the set of centroids obtained at the end of the $m$-th step of the RP$K$M. The std.error is defined as $\rho(m)=\frac{E_{m}^{*}-E_{m}}{E_{m}^{*}}$, where $E_{m}$ is the error for RP$K$M at the $m$-th step, and $E_{m}^{*}$ is the error obtained by $K$-means algorithm over the full dataset $D$, taking as initialization the centroids obtained by RP$K$M at the $m$-th step. Observe that $\rho(m)\leq 0$ and it measures the percentage of error with respect to the $K$-means over the entire dataset.

\vspace{-0.6cm}
In most of the cases, we observe a monotone descent of the centroid error with respect to the full dataset until convergence to the error associated to a solution of the $K$-means algorithm. This is remarkable since the approximation is constructed over a reduced number of representatives with respect to the total number of instances (see Fig.\ref{figu4}). In particular, at the third RP$K$M step, the error with respect to the $K$-means solution is under $10\%$. Evidently, as we increase the dimension, this percentage decreases until it achieves an error percentage smaller than $10\%$ and $5\%$ for the first and second RP$K$M step, respectively. 
Moreover, for $d=2$ and $K=9$, we observe no result for RP$K$M $1$, this is due to the fact that, in this case, the number of representatives is smaller than the number of clusters, i.e., $|\mathcal{P}_{1}|<K$. 

\subsubsection{Relation distance computations - quality of the approximation}
In this section, we fuse the results obtained at the previous sections and analyze, for the different algorithms, the trade-off between the number of distances computed and the quality of the obtained solutions.

\begin{figure}[H]
 \begin{center}
        \includegraphics[height=0.60\textheight,width=\textwidth]{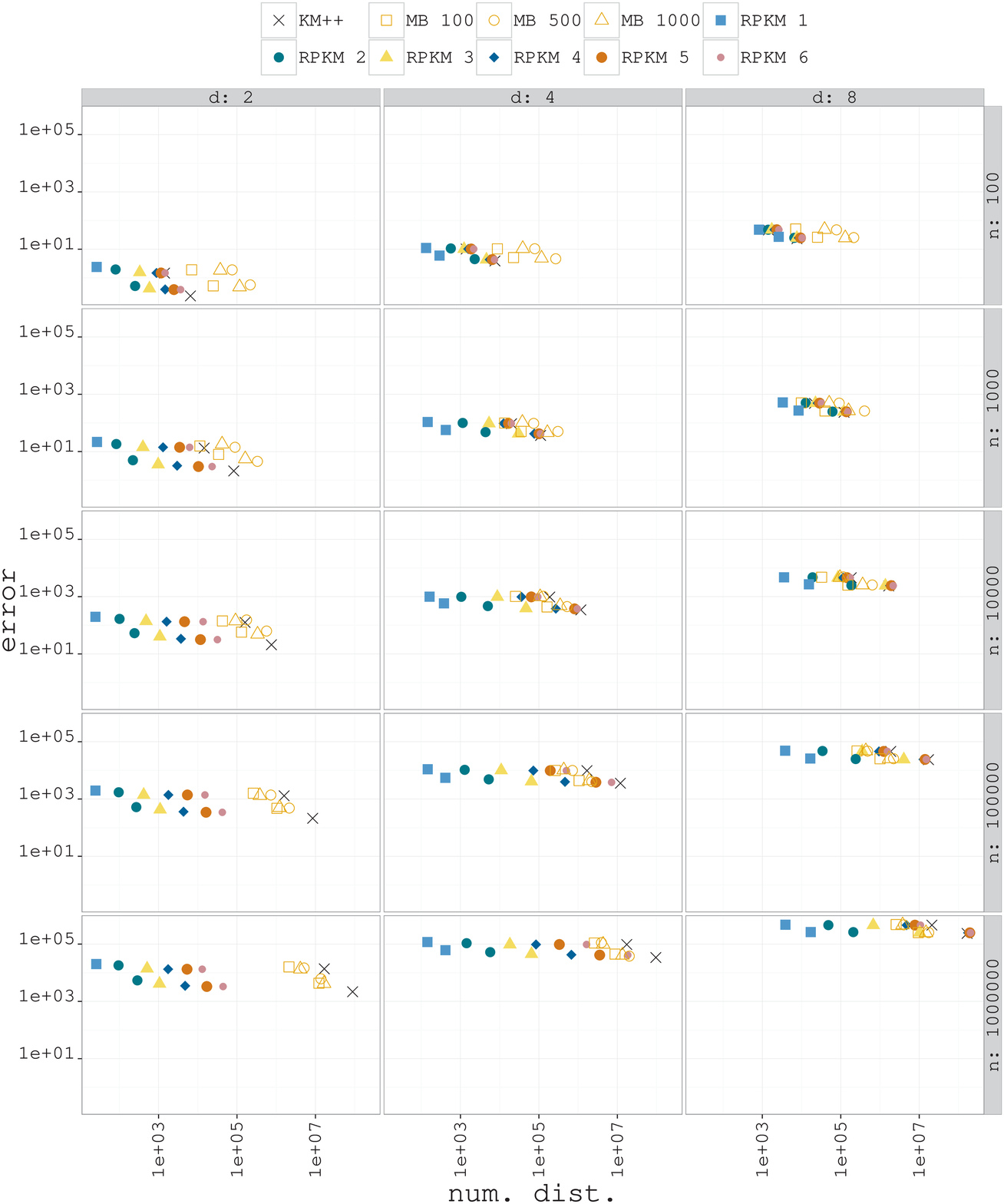}
        \caption{Quality of the approximation vs number of distance computations}
        \label{figu6}
         \end{center}
        \end{figure}

In Fig.\ref{figu6}, we show the relation between the number of distance computations and the error of the obtained solutions for RP$K$M, $K$M++ and MB.

Besides the dimensionality of the problem, as we increase the number of instances, the cloud of points associated to $K$M++ and MB separates
from the ones associated to the RP$K$M. This means that, as we increase the number of instances,
$K$M++ and MB require a much larger number of distance
computations in order to achieve a solution of similar quality to those obtained by the RP$K$M. In the best case scenario ($n=1000000$, $d=2$), RP$K$M reduces, at least, $6$ and $4$ orders of magnitude with respect to the $K$-means++ and the minibatch $K$-means. 
Evidently, for larger dimensions, the clouds associated to the RP$K$M, for the latter stages, can overlap those of $K$M++ and MB. 
In this case, even for the largest number of instances, we did not need to execute all the RP$K$M steps: at the fifth RP$K$M step, we have
already generated, approximately, as many representatives as instances.

In particular, consider the extreme cases: $d=2$, $n=1000000$ (case $1$) and $d=8$, $n=100$ (case $2$). In the first case, after the third RP$K$M step the standard error is already 
under $5\%$ and, after fourth step, the error is practically null. Such approximations are obtained after computing under $10^{-3}\%$ ($10^{-2}\%$ ) 
and slightly over $10^{-3}\%$ ($10^{-2}\%$ ) of the distances calculated by $K$M++ (MB) for RP$K$M $3$ and RP$K$M $4$, respectively.

In the second case, already after the first RP$K$M step, the standard error is under $8\%$ and, after the second step, the error is fairly close to zero. Such approximations 
are obtained after computing under $10^{-1}\%$ ($1\%$ )and under $1\%$ ($10\%$ ) of the distances calculated by $K$M++ (MB) for RP$K$M $1$ and RP$K$M $2$, respectively.

In the case of lower dimensions and greater number of instances, we require more RP$K$M steps to achieve an approximation with a similar standard error with 
respect to the case with greater dimensions and lower number of instances. As previously mentioned, this is due to the exponential growth of the number of representatives 
with respect to the dimension of the problem. Moreover, in the second case, we have a lower number of instances, hence, we need fewer RP$K$M steps in order to generate as many 
representatives as instances (in this example, this occurs for $m=3$). However, having a lower proportion of representatives with respect to the number of instances 
also implies a greater reduction in the number of distance computations with respect to the full dataset for the first case, while obtaining a similar standard error
in comparison to the second case.

\subsection{Real datasets}
In addition to the previous experimentation, we evaluate the performance of the grid based RP$K$M algorithm, $K$M++ and 
MB on a real-world dataset: the gas sensor array under dynamic gas mixtures dataset, which contains the 
acquired time series from $16$ chemical sensors exposed to gas mixtures at varying concentration levels \cite{Fonollosa}. 
The dataset consists of $4178504$ instances and $19$ attributes and is available in the UC-Irvine Machine Learning Repository.
The same experiment was performed over different datasets from the UC-Irvine Machine Learning Repository, achieving similar conclusions. For the
sake of brevity, the corresponding graphics are not included in this work.

Using this real-world dataset, we generate different subsamplings that we use to analyze the features of the algorithms. In particular, we take
$d \in \{2,4,8\}$ random attributes and $n \in \{4000,12000,40000,120000,400000,$ \\
$1200000,4000000\}$ random instances. The number
of clusters is $K \in \{3,9\}$. For each setting, we generate $10$ replicates
of the dataset.

\subsubsection{Distance computations}

For the real dataset experimentation, we perform the same analysis as that carried out for the artificial datasets case. 

\begin{figure}[H]
 \begin{center}
        \includegraphics[height=0.62\textheight,width=\textwidth]{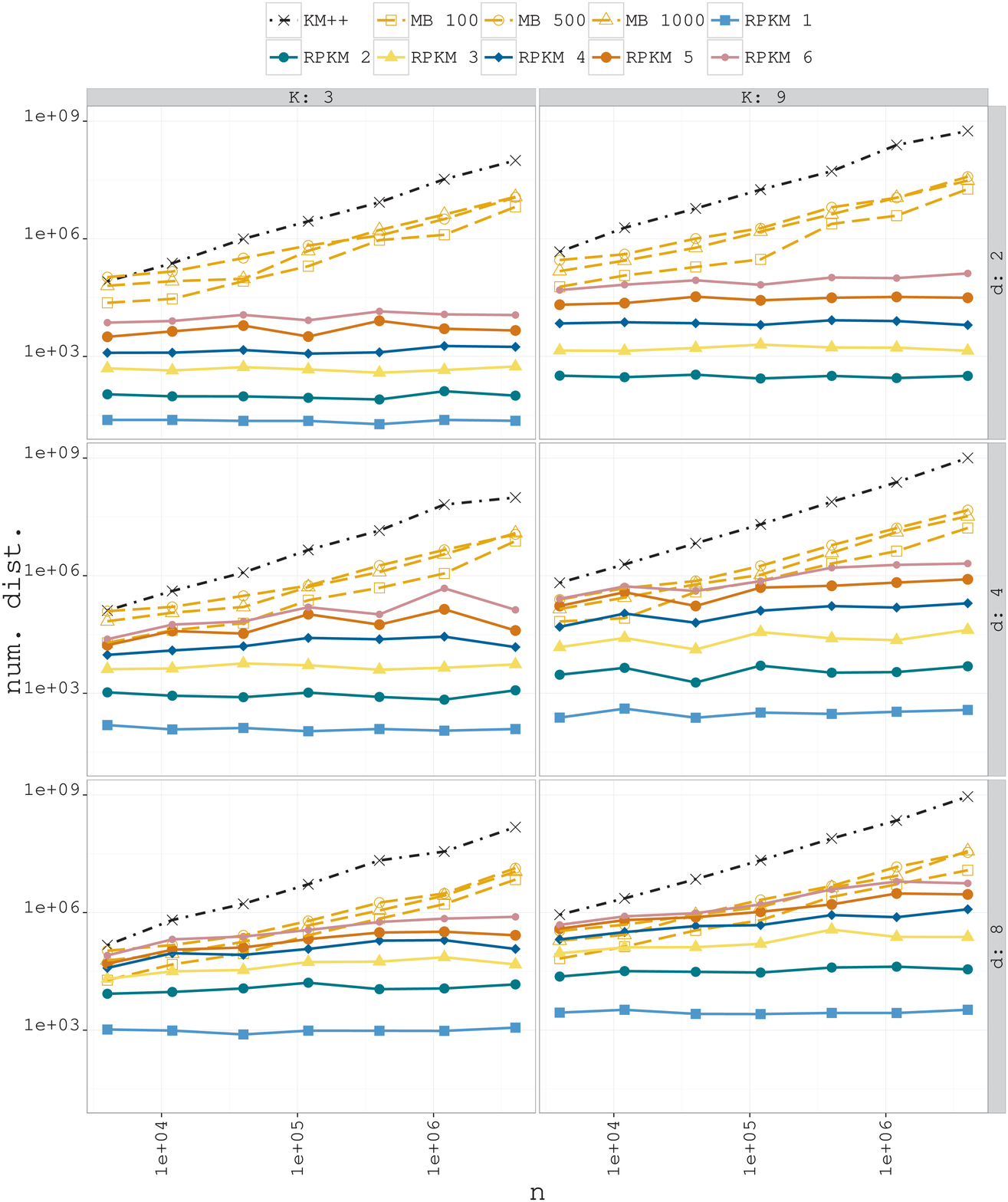}
        \caption{ Number of distance computations with respect to dataset size, $n$}
        \label{figu7}
         \end{center}
        \end{figure}

In  Fig.\ref{figu7}, as in Fig.\ref{figu3}, we present the relation between the number of distance computations and the dataset size, this time for the real dataset.
In general, we observe a similar behavior with respect to the artificial datasets case: RP$K$M reduces, in many orders of magnitude, the number of distance computations with respect
to $K$M++ and MB. However, in this case, even at the last step of RP$K$M, the number of distances does not always increase with respect to the number of instances.
 
\begin{figure}[H]
 \begin{center}
        \includegraphics[height=0.5\textheight,width=\textwidth]{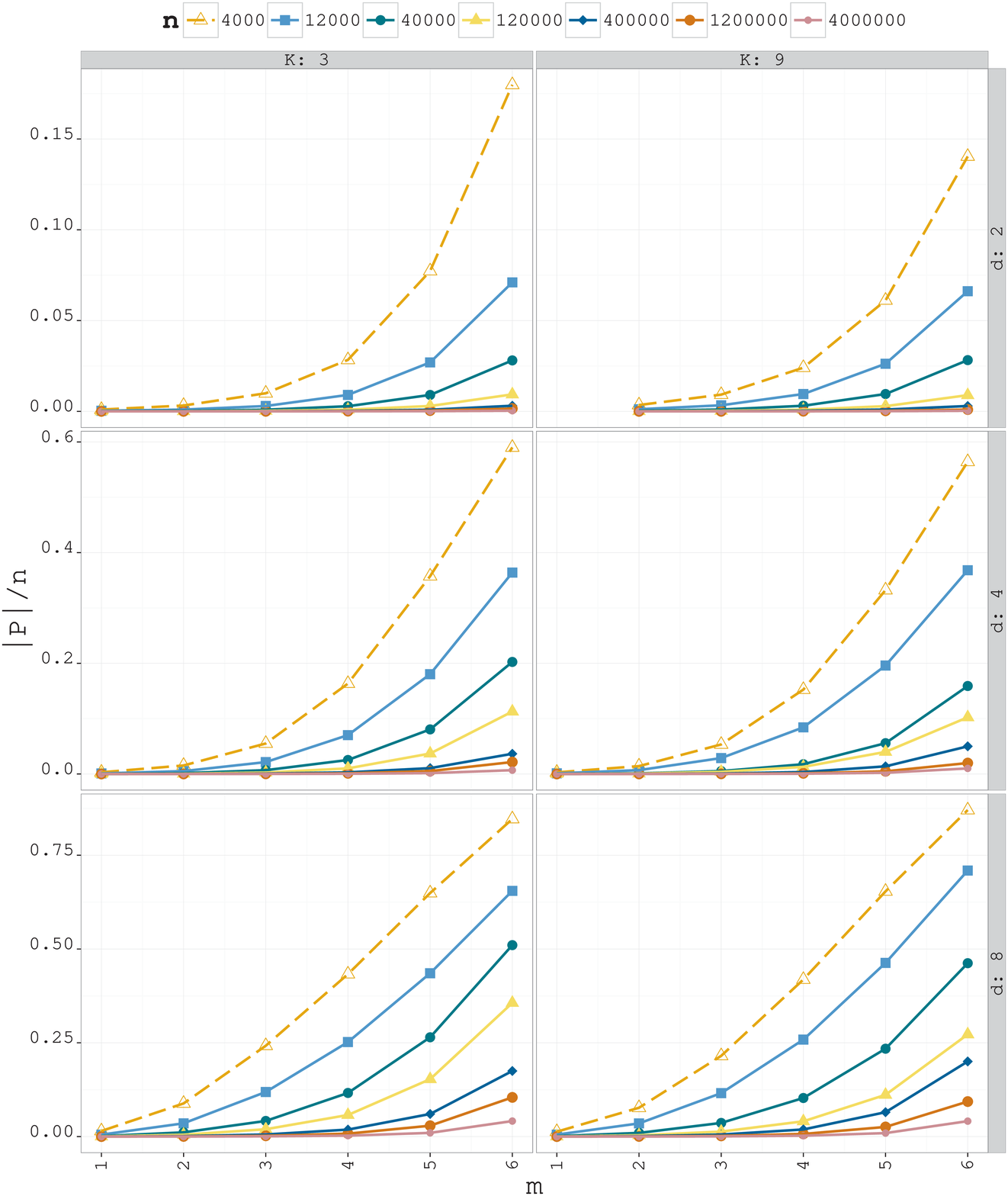}
     \vspace{-0.2cm}
        \caption{ Percentage of subsets with respect to RP$K$M step}
        \label{figu8}
         \end{center}
        \end{figure}

In particular, as can be verified in Fig.\ref{figu8}, the number of representatives does not necessarily increase as we consider higher dataset sizes. This is due to the fact that the data points, in this case,
are grouped into more condensed clouds than in the case of the artificial Gaussian dataset case. Hence, it is plausible to observe that the number of distance computations, even at the latter stages of the RP$K$M, does not necessarily increase with respect to the dataset 
size. Furthermore, for low dimensions, this fact implies that, even at the last RP$K$M step, we have a lower number of distance computations than any version of MB and $K$M++. 

\vspace{-0.6cm}
In more detail, we notice that for the largest number of instances, the last step of the RP$K$M executes less than $1\%$ and $0.1\%$ of the distances computed by  
MB and $K$M++, respectively. For the largest dimension considered ($d=8$), the latter steps RP$K$M execute a similar order of distance computations with respect to  
MB. However, intermediate RP$K$M steps (RP$K$M $3$) still computes less than $1\%$ and $0.1\%$ of the distances computed with respect to
MB and $K$M++. It is important to remember that the number of distance computations for $K$M++ and MB is 
independent of the dimensionality of the problem, which is not usually true for the RP$K$M algorithm.

\vspace{-0.6cm}
For the analyzed dataset, we observe that the number of representatives does not grow as rapidly as in the artificial sets case. In particular, for the lowest 
dimension, the number of representatives barely achieves $20\%$ of the dataset size at the last grid based RP$K$M step. As we increase the dimensionality, and therefore
generate more representatives, this number grows to over $75\%$ for the shortest dataset size case.

\subsubsection{Quality of the approximation}

\begin{figure}[H]
 \begin{center}
        \includegraphics[height=0.47\textheight,width=\textwidth]{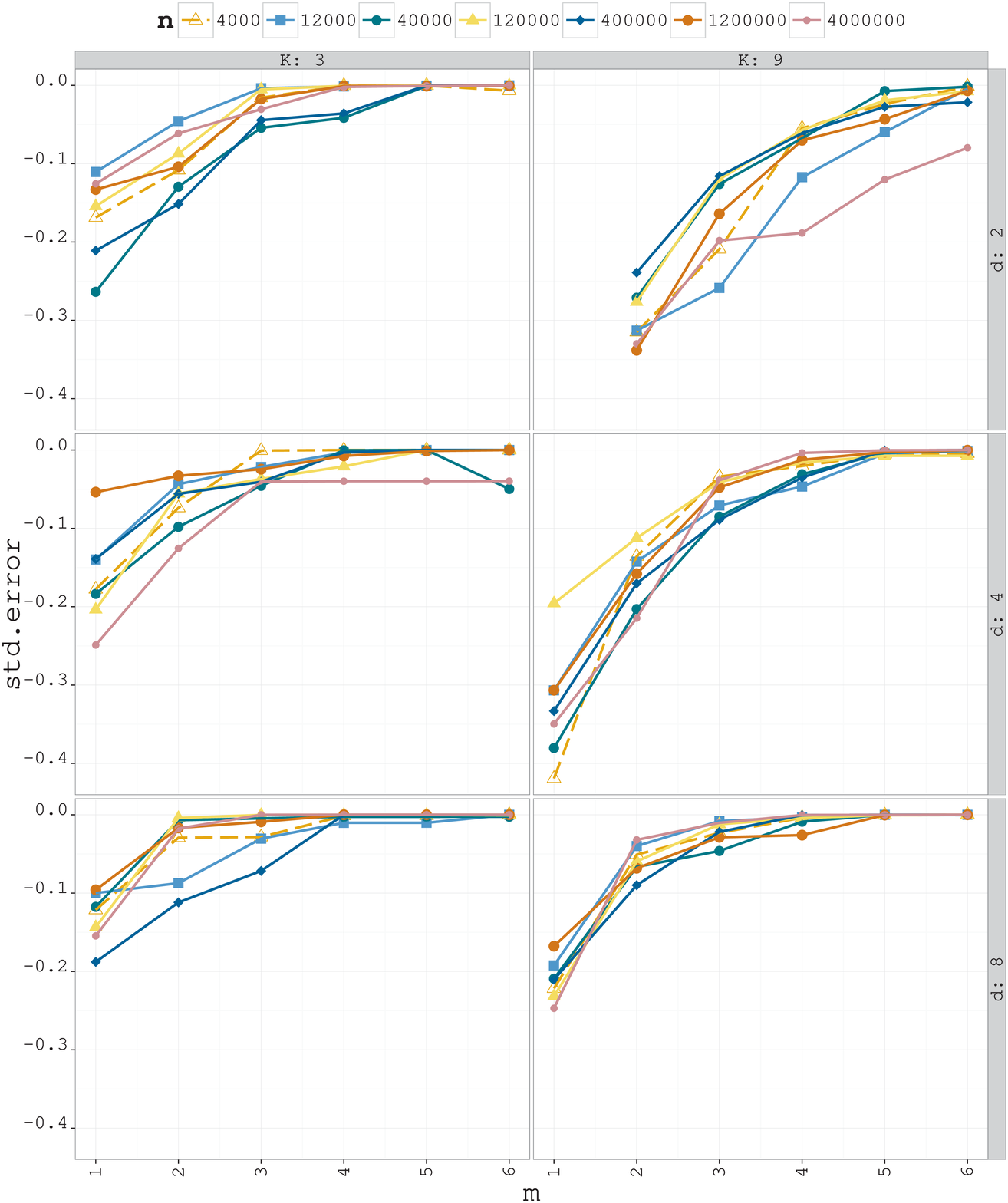}
        \caption{Quality of the approximation with respect to RP$K$M step}
        \label{figu9}
         \end{center}
        \end{figure}

In Fig.\ref{figu9}, we observe the evolution of the standarized error, for the full dataset, with respect to the set of centroids obtained at the $m$-th
step of the RP$K$M. As commented in the artificial datasets case, in most of the cases, there is a monotone descent of the centroid error with respect to the full dataset until convergence to the error associated to a solution of the $K$-means algorithm over the full dataset. Commonly, at the third RP$K$M step, the such standarized error is under $10\%$. This is remarkable since, as can be seen in Fig.\ref{figu7}, the number of distances computed by RP$K$M $3$, compared to MB and $K$M++, is under 
$10^{-3} \ \%$ and $10^{-4} \ \%$, respectively. 

\vspace{-0.6cm}

As we increase the dimension the standarized error, in most of the cases, is under $10\%$, even for RP$K$M $2$ on the largest dataset size case. 

\subsubsection{Relation distance computations - quality of the approximation}

\begin{figure}[H]
 \begin{center}
        \includegraphics[height=0.59\textheight,width=\textwidth]{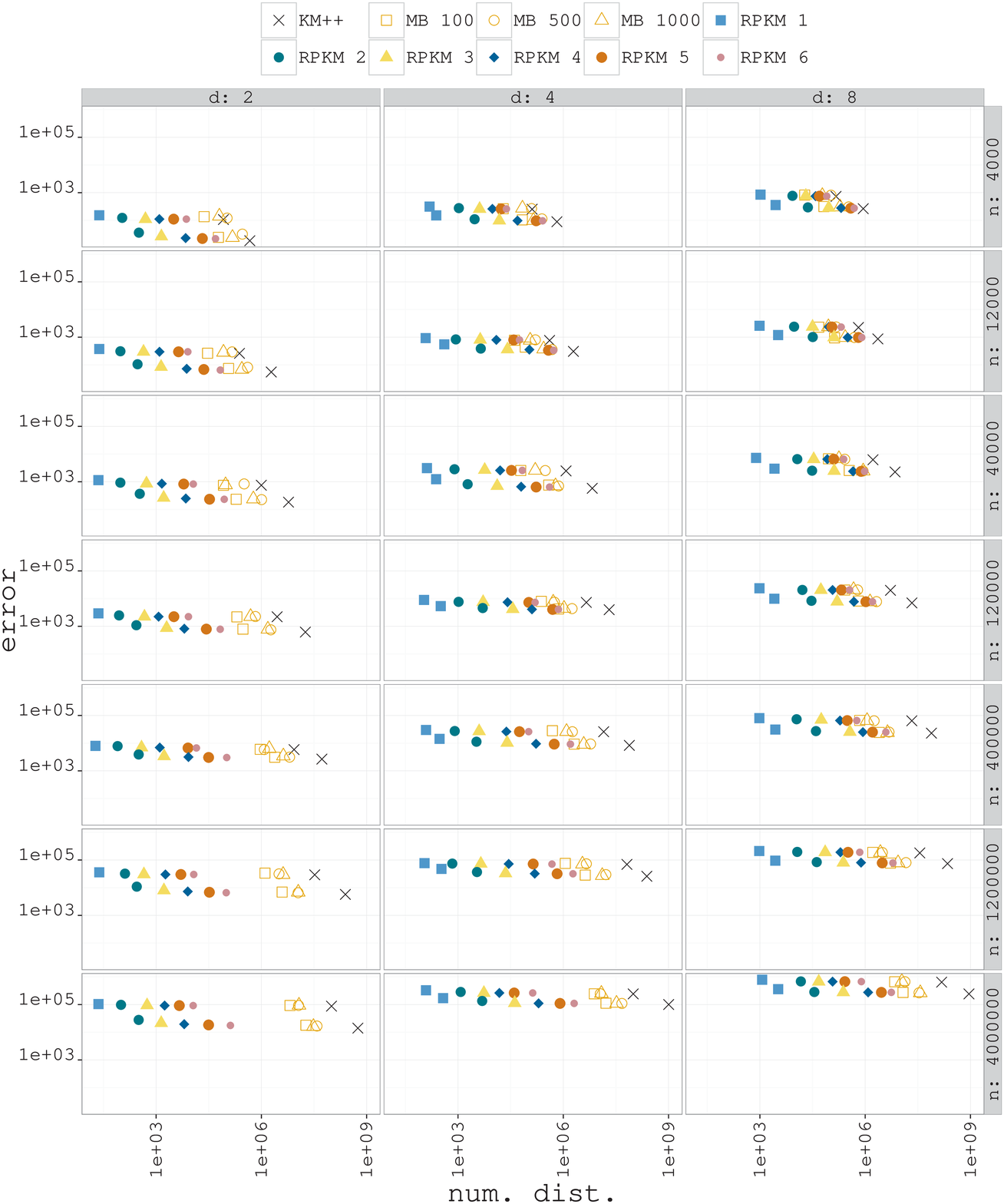}
        \caption{Quality of the approximation vs number of distance computations}
        \label{figu10}
         \end{center}
        \end{figure}

In order to have a better understanding of the relation distance computations against overall error, we consider Fig.\ref{figu10}.
As we observed in the artificial dataset case, when we increase the number of instances, the cloud of points associated to $K$M++ and MB 
separates from the ones associated to the RP$K$M. As we increase the number of instances, $K$M++ and MB require a much larger
number of distance computations in order to achieve a solution of similar quality than those obtained by the RP$K$M. For the lowest dimensional case, even for an 
intermediate RP$K$M step, it reduces over $4$ orders $(d=2,n=4000000)$ and $5$ orders of magnitude $(d=2,n=4000000)$ with respect to MB and $K$M++, respectively. 
Evidently, as we increase the dimensionality of the problem and,
therefore, generate a larger amount of representatives and compute more distances, we still observe a reduction of 
$4$ and $5$ orders of magnitude, but for the first RP$K$M step. In addition, we can see that the minibatch k-means with the smallest batch, MB $100$,
has, in some cases, a similar amount of distance computations with respect to the second step of RP$K$M.

Such a reduction in the number of distance computations is achieved while still obtaining competitive approximations. For instance, even in the case of greater dimension and lower 
number of instances ($d=8$, $n=4000$), after the second RP$K$M step the standard error is already  under $5\%$ and, after the third step, the error is practically null. 
These approximations are obtained after computing under $7\cdot 10^{-4}\%$ ($2\cdot 10^{-3}\%$ )and $10^{-3}\%$ ($5\cdot 10^{-3}\%$ )
of the distances calculated by $K$M++ (MB), for RP$K$M $3$ and RP$K$M $4$, respectively.

\vspace{-0.6cm}

In general, we observe that the grid based RP$K$M algorithm is able to generate competitive approximations with a significant reduction
in the number of distance computations. As the dataset size increases, we observed a more drastic reduction in the number of distance computations as
can be seen in Fig.\ref{figu6} and Fig.\ref{figu10}. We observe the same behavior as we increase the dimension of the instances, 
however, the order of reduction with respect to MB and $K$M++ decreases, as expected. This is due to the fact that the number of RP$K$M representatives,
in this case, grows exponentially with respect to the dimension of the dataset.

\section{Final comments and future work}

In this work, we present an alternative to the $K$-means algorithm applicable to massive data problems called recursive partition based $K$-means (RP$K$M).
This approach recursively partitions the entire dataset into a small number of subsets, each of which is characterized by its representative (center of mass) and weight
(cardinality), after that a weighted version of the Lloyd's algorithm is applied over this local representation. The objective is to describe the full dataset by this
representation, which ultimately leads to a reduction of distance computations. Indeed, in the experimental section, we observe that the RP$K$M algorithm generates
competitive approximations, even at its earlier iterations, while reducing several orders of magnitude of distance computations.

In Section $3$, we have derived some theoretical properties of the RP$K$M. Among
other results, we can guarantee the non repetition of the clusterings generated at each RP$K$M iteration (except the last one), which ultimately implies the reduction of
the total amount of Lloyd iterations, as well as leading, in most of the cases, to a monotone decrease of the overall error function.

In the experimental section, the grid based RP$K$M was compared with two well-known approaches: the $K$-means++ and minibatch $K$-means. In this analysis, we observed a 
dramatic reduction in the number of distance computations with respect to both of them, as well as a consistent monotone decrease of the error function.
Since the RP$K$M algorithm seeks to reduce the number of representatives used per iteration, we observed a larger reduction in the number of distance computations
as we enlarged the number of instances of the dataset. Furthermore, at the earlier stages of the RP$K$M, the size of the dataset did not have a relevant impact on the
number of iterations or distance computations for the associated weighted $K$-means problem. Thus, the number of computations, especially for massive data applications, 
can be greatly reduced.

On the other hand, it is important to remark that the number of subpartitions generated at each iteration of the grid based RP$K$M grows exponentially with respect to the dimension, $d$, of the dataset.
As we can see in Fig.\ref{figu1},
some of these subpartitions might have a small probability of changing their current cluster affiliation (with respect to the assignment given by the previous RP$K$M iteration).
In this sense, it might be more valuable to partition the  areas that are more likely to have subpartitions associated with different clusters.

One possible approach consists of characterizing the subsets that lie on a cluster boundary, i.e., subsets that are close to two or more 
clusters \cite{Hung}. In this approach, the number of representatives does not grow exponentially with respect to the dimension of the dataset. 
For this reason, as a future step, we plan to define a low computational cost algorithm to  determine the cluster boundary at each iteration. 
The subsets in this area will have a greater priority when selecting the regions that we want to partition in the next RP$K$M iteration.

One last, but still important, advantage of the RP$K$M algorithm is the fact that its parallelization is direct. The RP$K$M algorithm mainly depends on two steps: The data 
partition process and the application of the weighted version of Lloyd's algorithm. In the first step, each point can independently decide which 
subset it belongs to, hence the construction of the set of representatives and weights can be done in a parallel manner. Analogously, for the second phase, given a set 
of prototypes, each data point can separately decide which cluster it belongs to and the update of the centroid can be simply computed by averaging the
points \cite{Dean,Zhao}. For this reason, we also plan to implement the RP$K$M algorithm on a parallel framework such as {\it Apache Spark}.

\section*{Acknowledgments}

Marco Cap\'o and Aritz P\'erez are partially supported by the Basque Government, Elkartek and by the Spanish Ministry of Economy and Competitiveness MINECO: BCAM Severo Ochoa excelence accreditation SVP-2014-068574 and SEV-2013-0323. Jos\'e A. Lozano  is partially supported by the Basque Government (IT609-13), Elkartek and the Spanish Ministry of Economy and Competitiveness MINECO (TIN2013-41272P).

\appendix
\section{RP$K$M properties} \label{App:AppendixA}

\primelemma*
\begin{proof}

First of all observe that $|D|=\sum_{R \in \mathcal{P}} |R|$ and $\overline{D}= \frac{\sum_{R \in \mathcal{P}} |R|\cdot\overline{R}}{|D|}$. Given any pair $\textbf{c}$, $\textbf{c}^{'} \in \mathbb{R}^d$, we have that
\begin{eqnarray*}
f(\textbf{c})&=& |D|\cdot\| \overline{D}- \textbf{c} \|^2 - \sum_{R \in \mathcal{P}} |R|\cdot\| \overline{R}- \textbf{c} \|^2 \\
&=& |D|\cdot(\| \overline{D}- \textbf{c}^{'} \|^2 + 2(\textbf{c}^{'}-\textbf{c})^{t}(\overline{D}-\textbf{c}^{'})+\| \textbf{c}^{'}- \textbf{c} \|^2)\\ 
& & -\sum_{R \in \mathcal{P}} |R|\cdot(\| \overline{R}- \textbf{c}^{'} \|^2 + 2(\textbf{c}^{'}-\textbf{c})^{t}(\overline{R}-\textbf{c}^{'})+\| \textbf{c}^{'}- \textbf{c} \|^2) \\
&=& f(\textbf{c}^{'})+ 2|D|(\textbf{c}^{'}-\textbf{c})^{t}(\overline{D}-\textbf{c}^{'})- 2\sum_{R \in \mathcal{P}} |R|(\textbf{c}^{'}-\textbf{c})^{t}(\overline{R}-\textbf{c}^{'}) \\
&=& f(\textbf{c}^{'})
\end{eqnarray*}
\end{proof}
\seclemma*

\begin{proof}
Let the index of $S \in \mathcal{P}$ on the cluster $\mathcal{G}=\{G_{k}\}_{k=1}^K$, $i(S,\mathcal{G})$, be defined as 
$i(S,\mathcal{G})=\{k\ |\ S \subseteq G_{k} \}$, then
\begin{eqnarray*}
E_{\mathcal{P}}(\mathcal{G})-E_{\mathcal{P}^{'}}(\mathcal{G})&=&\sum_{S \in \mathcal{P}} |S|\cdot\| \overline{S}-\overline{G_{i(S,\mathcal{G})}} \|^2 - \sum_{S^{'} \in \mathcal{P}^{'}} |S^{'}|\cdot\| \overline{S^{'}}-\overline{G_{i(S^{'},\mathcal{G})}}\|^2 \\
&=& \sum_{S \in \mathcal{P}} (|S|\cdot\| \overline{S}-\overline{G_{i(S,\mathcal{G})}} \|^2 - \sum_{R \in \mathcal{P}^{'}[S]} |R|\cdot\| \overline{R}-\overline{G_{i(R,\mathcal{G})}} \|^2 ) \\
&=& \sum_{S \in \mathcal{P}} (|S|\cdot\| \overline{S}-\overline{G_{i(S,\mathcal{G})}} \|^2 - \sum_{R \in \mathcal{P}^{'}[S]} |R|\cdot\| \overline{R}-\overline{G_{i(S,\mathcal{G})}} \|^2 ) \\
&=& \sum_{S \in \mathcal{P}} (|S|\cdot\| \overline{S}-\overline{G_{i(S,\mathcal{G}^{'})}^{'}} \|^2 - \sum_{R \in \mathcal{P}^{'}[S]} |R|\cdot\| \overline{R}-\overline{G_{i(S,\mathcal{G}^{'})}^{'}} \|^{2} ) \\
&=& \sum_{S \in \mathcal{P}} (|S|\cdot\| \overline{S}-\overline{G_{i(S,\mathcal{G}^{'})}^{'}} \|^2 - \sum_{R \in \mathcal{P}^{'}[S]} |R|\cdot\| \overline{R}-\overline{G_{i(R,\mathcal{G}^{'})}^{'}} \|^2 ) \\
&=&\sum_{S \in \mathcal{P}} |S|\cdot\| \overline{S}-\overline{G_{i(S,\mathcal{G}^{'})}^{'}} \|^2 - \sum_{S^{'} \in \mathcal{P}^{'}} |S^{'}|\cdot\| \overline{S^{'}}-\overline{G_{i(S^{'},\mathcal{G}^{'})}^{'}}\|^2 \\
&=& E_{\mathcal{P}}(\mathcal{G}^{'})-E_{\mathcal{P}^{'}}(\mathcal{G}^{'}) \\
\end{eqnarray*}
\end{proof}
Before prooving Corollary \ref{thm:monotonedescend} and Lemma \ref{lemma:norepetitions}, we analyze the error of a cluster with respect to a weighted $K$-means iteration. 
We observe that following inequality is satisfied at the $r$-th weighted Lloyd's algorithm iteration:

\begin{eqnarray}\label{eccGWK}
E_{\mathcal{P}}(C_{r})\geq E_{\mathcal{P}}(\mathcal{G}_r) \geq E_{\mathcal{P}}(C_{r+1})
\end{eqnarray}

Furthermore, observe that, if $E_{\mathcal{P}}(C_{r})= E_{\mathcal{P}}(\mathcal{G}_r) $, then, after the update step of the weighted Lloyd's algorithm, we obtain $C_{r}=C_{r+1}$ and the algorithm stops at the ($r+1$)-th iteration. On the other hand, if $E_{\mathcal{P}}(C_{r}) > E_{\mathcal{P}}(\mathcal{G}_r) = E_{\mathcal{P}}(C_{r+1})$, then, in the following weighted Lloyd's algorithm iteration, we obtain 
$E_{\mathcal{P}}(C_{r+1})=E_{\mathcal{P}}(\mathcal{G}_{r+1})=E_{\mathcal{P}}(C_{r+2})$ and $C_{r+1}=C_{r+2}$, hence the algorithm
stops, at most, at the ($r+2$)-th iteration.

Hence, we have the following chain of inequalities for any weighted Lloyd's algorithm run 

\begin{eqnarray}\label{eccCI}
E_{\mathcal{P}}(C_{0})>E_{\mathcal{P}}(\mathcal{G}_0)> E_{\mathcal{P}}(C_{1})> E_{\mathcal{P}}(\mathcal{G}_1)> E_{\mathcal{P}}(C_{2})>\nonumber\\
\ldots > E_{\mathcal{P}}(\mathcal{G}_{l-2}) \geq E_{\mathcal{P}}(C_{l-1}) \geq E_{\mathcal{P}}(\mathcal{G}_{l-1}) \geq E_{\mathcal{P}}(C_{l}), 
\end{eqnarray}

where $l$ is the total number of weighted Lloyd iterations.

\sectheo*
\begin{proof}
From Lemma \ref{lemma:invariance} and Eq.\ref{eccCI}, we have the following inequalities:

\begin{eqnarray}\label{eccx7}
E_{\mathcal{P}_{i}}(\mathcal{G}_{l_{i-1}-1}^{i-1})\geq E_{\mathcal{P}_{i}}(C_{i-1})\geq E_{\mathcal{P}_{i}}(\mathcal{G}_{l_{i}-1}^{i}) \geq E_{\mathcal{P}_{i}}(C_{i})
\end{eqnarray}
\vspace{-0.90cm}
\begin{eqnarray}\label{eccx8}
E_{\mathcal{P}_{i}}(\mathcal{G}_{l_{i-1}-1}^{i-1})-E_{\mathcal{P}_{i}}(\mathcal{G}_{l_{i}-1}^{i})=E_{\mathcal{P}_{j}}(\mathcal{G}_{l_{i-1}-1}^{i-1})-E_{\mathcal{P}_{j}}(\mathcal{G}_{l_{i}-1}^{i})=\xi_i \geq 0
\end{eqnarray}
 
Eq.\ref{eccx7} holds for $i \in \{1,\ldots, m\}$ and Eq.\ref{eccx8} for any $j>i$. From Eq.\ref{eccx8}, we can see that the difference between both clusterings remain constant for any thinner partition $\mathcal{P}_{j}$. A consequence of this is that, if we take $\mathcal{P}_{j}$ as a partition thin enough such that there is only one instance per subset, then 
adding the difference clustering error (associated to both centroids) for the partitions $\mathcal{P}_j$ and $\mathcal{P}_i$, we have the following relation over the error for the entire dataset (observe that the following relation holds in general for any partition thinner than $\mathcal{P}_{i}$):

\begin{eqnarray*}
E(C_{i}) \leq E(C_{i-1}) \Longleftrightarrow E(\mathcal{G}_{l_{i-1}-1}^{i-1})- E(C_{i-1}) \leq \xi_{i} + (E(\mathcal{G}_{l_{i}-1}^{i})- E(C_{i}))
\end{eqnarray*}

\end{proof}

\terlemma*
\begin{proof}
Using the chain of inequalities (\ref{eccCI}), we observe that the following inequalities hold at the first iteration of the RP$K$M:

\begin{eqnarray}\label{eccx5}
E_{\mathcal{P}_{1}}(C_{0}^{1}=C_{0})>E_{\mathcal{P}_{1}}(\mathcal{G}_0^{1})> E_{\mathcal{P}_{1}}(C_{1}^{1})> E_{\mathcal{P}_{1}}(\mathcal{G}_1^{1})>E_{\mathcal{P}_{1}}(C_{2}^{1})>\nonumber\\
\ldots > E_{\mathcal{P}_1}(\mathcal{G}_{a_1-2}^{1}) \geq E_{\mathcal{P}_{1}}(C_{l_1-1}^{1}) \geq E_{\mathcal{P}_{1}}(\mathcal{G}_{l_1-1}^{1})\geq E_{\mathcal{P}_{1}}(C_1=C_{l_1}^{1}) 
\end{eqnarray}

Analogously, for the subsequent $i$-th iteration of the RP$K$M algorithm, we obtain the  following set of inequalities

\begin{eqnarray}\label{eccx6}
E_{\mathcal{P}_{i}}(\mathcal{G}_{a_{i-1}-1}^{i-1})>E_{\mathcal{P}_{i}}(C_{0}^{i}=C_{i-1})>E_{\mathcal{P}_{i}}(\mathcal{G}_0^{i})> E_{\mathcal{P}_{i}}(C_{1}^{i})> E_{\mathcal{P}_{i}}(\mathcal{G}_1^{i})>\nonumber\\
E_{\mathcal{P}_{i}}(C_{2}^{i})> \ldots > E_{\mathcal{P}_i}(\mathcal{G}_{l_i-2}^{i}) \geq E_{\mathcal{P}_{i}}(C_{l_i-1}^{i}) \geq E_{\mathcal{P}_{i}}(\mathcal{G}_{l_i-1}^{i})\nonumber\\
 \geq E_{\mathcal{P}_{i}}(C_i=C_{l_i}^{i}), \ i \in \{2,\ldots,m\}
\end{eqnarray}

First of all, observe that the error associated to all the clusters generated at the $j$-th RP$K$M iteration keep the same ordering for the
error associated to a thinner partition $\mathcal{P}_i$. In particular, we have $E_{\mathcal{P}_{i}}(\mathcal{G}_{l_j-1}^{j}) \leq E_{\mathcal{P}_{i}}(\mathcal{G}_{s}^{j})$ for  $s< l_{j}-1$.
To verify this we make use of Lemma \ref{lemma:invariance}, from which we know that  $E_{\mathcal{P}_{i}}(\mathcal{G}_{l_j-1}^{j})-E_{\mathcal{P}_{i}}(\mathcal{G}_{s}^{j})=E_{\mathcal{P}_{j}}(\mathcal{G}_{l_j-1}^{j})-E_{\mathcal{P}_{j}}(\mathcal{G}_{s}^{j}) \leq 0 \rightarrow E_{\mathcal{P}_{i}}(\mathcal{G}_{l_j-1}^{j}) \leq E_{\mathcal{P}_{i}}(\mathcal{G}_{s}^{j})$ for $s< l_{j}-1$.
This means that, the last clustering found at the $j$-th RP$K$M iteration also has the smallest error, with the partition $\mathcal{P}_i$, with respect to the
previous clusters obtained at the $j$-th RP$K$M iteration.

From the chain of inequalities (\ref{eccx5}) and (\ref{eccx6}), we know that 

\begin{eqnarray}\label{ecc5}
E_{\mathcal{P}_{h}}(\mathcal{G}_{l_{h-1}-1}^{h-1})\geq E_{\mathcal{P}_{h}}(\mathcal{G}_{l_{h}-1}^{h}) \ \forall h \in \{j+1,\ldots, i-1\}, 
\end{eqnarray}

where, if the equality holds, then $a_{h}=1$. From Lemma \ref{lemma:invariance}, ($\ref{ecc5}$) implies

\begin{eqnarray}\label{ecc6}
E_{\mathcal{P}_{i}}(\mathcal{G}_{l_{j}-1}^{j})\geq E_{\mathcal{P}_{i}}(\mathcal{G}_{l_{j+1}-1}^{j+1})\geq \ldots \geq E_{\mathcal{P}_{i}}(\mathcal{G}_{l_{i-1}-1}^{i-1})
\end{eqnarray}

In other words, the error with respect to the current partition (or any thinner partition) of the optimal patterns obtained at each RP$K$M iteration 
decreases monotonically. 

By {\it reductio ad absurdum}, if we assume that $\mathcal{G}_{r}^{i}=\mathcal{G}_{s}^{j} $, for some $r \in \{1,\ldots, l_{i}-1\}$ and 
$s \in \{1,\ldots, l_{j}-1\}$ and that there exists $j<h<i$ such that $l_{h}>1$, then $E_{\mathcal{P}_{i}}(\mathcal{G}_{l_{j}-1}^{j})>E_{\mathcal{P}_{i}}(\mathcal{G}_{l_{i-1}-1}^{i-1}) \Rightarrow E_{\mathcal{P}_{i}}(\mathcal{G}_{s}^{j})>E_{\mathcal{P}_{i}}(\mathcal{G}_{r}^{i})=E_{\mathcal{P}_{i}}(\mathcal{G}_{s}^{j})$ ($\Rightarrow\Leftarrow$).
Therefore, from now on we assume $l_{j+1}=\ldots=l_{i-1} = 1$.

Analogously, if we assume that $\mathcal{G}_{r}^{i}=\mathcal{G}_{s}^{j} $, for some $r \in \{1,\ldots, l_{i}-1\}$ and 
$s \in \{1,\ldots, l_{j}-1\}$ and that $l_i > 1$, then $E_{\mathcal{P}_{i}}(\mathcal{G}_{l_{j}-1}^{j}) \geq E_{\mathcal{P}_{i}}(\mathcal{G}_{l_{i-1}-1}^{i-1})>E_{\mathcal{P}_{i}}(\mathcal{G}_{r}^{i})=E_{\mathcal{P}_{i}}(\mathcal{G}_{s}^{j}) \Rightarrow E_{\mathcal{P}_{i}}(\mathcal{G}_{s}^{j})>E_{\mathcal{P}_{i}}(\mathcal{G}_{s}^{j})$ ($\Rightarrow\Leftarrow$). 

In the case that $l_{j+1}=\ldots=l_{i} = 1$,  the error associated to $\mathcal{G}_{r}^{i}$ satisfies $E_{\mathcal{P}_{i}}(\mathcal{G}_{s}^{j})\geq E_{\mathcal{P}_{i}}(\mathcal{G}_{l_{j}-1}^{j}) \geq E_{\mathcal{P}_{i}}(\mathcal{G}_{l_{i-1}-1}^{i-1})\geq E_{\mathcal{P}_{i}}(\mathcal{G}_{r}^{i})$, hence the only possible choice is
$s=l_j$.
\end{proof}

\primtheo*

\begin{proof}

Lemma \ref{lemma:norepetitions} implies that, at each RP$K$M iteration, no previous clustering can be repeated (except the last one from the clustering sequence).
Therefore, we can eliminate, at least, all clusters generated at the previous RP$K$M iterations except the last one ($\sum\limits_{j=1}^{i-1} (l_j-1)$). Moreover,
the number of different partitions for $|\mathcal{P}_{i}|$ observations into $K$ groups is a Stirling number of the second kind, ${|\mathcal{P}_{i}| \brace K}$ \cite{Sami}, then 
$l_i \leq {|\mathcal{P}_{i}| \brace K}-\sum\limits_{j=1}^{i-1} (l_j-1)$.

\end{proof}

\end{document}